\documentclass{nle}

\usepackage{amsmath}
\usepackage{graphicx}
\usepackage{natbib}
\ifpdf%
\usepackage{epstopdf}%
\else%
\fi
\usepackage{multirow}
\usepackage{caption}
\usepackage{longtable}
\usepackage{soul}

\setstcolor{blue}

\setstcolor{red}
\begin{document}
\label{firstpage}

\lefttitle{Mandal {\it et al}}
\righttitle{Natural Language Engineering}

\papertitle{Article}

\jnlPage{1}{00}
\jnlDoiYr{2019}
\doival{10.1017/xxxxx}

\title{Is Attention always needed? A Case Study on Language Identification from Speech}

\begin{authgrp}
\author{Atanu Mandal$^{1\star}$, Santanu Pal$^2$, Indranil Dutta$^3$, Mahidas Bhattacharya$^3$, and Sudip Kumar Naskar$^1$}
\affiliation{$^1$Department of Computer Science and Engineering, Jadavpur University, Kolkata \\ $^2$Wipro AI Lab, Wipro India Limited, Bengaluru \\ $^3$School of Languages and Linguistics, Jadavpur University, Kolkata \\
$^\star$Corresponding author. E-mails:
        \email{atanumandal0491@gmail.com, santanu.pal2@wipro.com, indranildutta.lnl@jadavpuruniversity.in, languagemahib@gmail.com, sudipkumar.naskar@jadavpuruniversity.in}\\Competing interests: The author(s) declare none}
\end{authgrp}

\history{(Received xx xxx xxx; revised xx xxx xxx; accepted xx xxx xxx)}

\begin{abstract}
Language Identification (LID) is a crucial preliminary process in the field of Automatic Speech Recognition (ASR) that involves the identification of a spoken language from audio samples. Contemporary systems that can process speech in multiple languages require users to expressly designate one or more languages prior to utilization. The LID task assumes a significant role in scenarios where ASR systems are unable to comprehend the spoken language in multilingual settings, leading to unsuccessful speech recognition outcomes. The present study introduces convolutional recurrent neural network (CRNN) based LID, designed to operate on the Mel-frequency Cepstral Coefficient (MFCC) characteristics of audio samples. Furthermore, we replicate certain state-of-the-art methodologies, specifically the Convolutional Neural Network (CNN) and Attention-based Convolutional Recurrent Neural Network (CRNN with attention), and conduct a comparative analysis with our CRNN-based approach. We conducted comprehensive evaluations on thirteen distinct Indian languages and our model resulted in over 98\% classification accuracy. The LID model exhibits high-performance levels ranging from 97\% to 100\% for languages that are linguistically similar. The proposed LID model exhibits a high degree of extensibility to additional languages and demonstrates a strong resistance to noise, achieving 91.2\% accuracy in a noisy setting when applied to a European Language (EU) dataset.
\end{abstract}

\maketitle

\section{Introduction}
\label{sec:Intro}
In the era of the Internet of Things, smart and intelligent assistants (e.g., Alexa \footnote{https://developer.amazon.com/en-US/alexa/alexa-voice-service}; Siri \footnote{https://www.apple.com/in/siri/}; Cortana \footnote{https://www.microsoft.com/en-in/windows/cortana}; Google Assistant \footnote{https://assistant.google.com/}; etc.) can interact with humans with some default language settings (mostly in English) and these smart assistants rely heavily on ASR. The motivation for our work stems from the inadequacy of virtual assistants in providing support in multilingual settings. In order to enhance the durability of intelligent assistants, LID can be implemented to enable automatic recognition of the speaker's language, thereby facilitating appropriate language setting adjustments. Psychological behaviour exhibits that Humans have an inherent capability to determine the language of a statement nearly instantly. Automatic LID seeks to classify a speaker's language usage from their speech utterances.

We focus our study of LID on Indian Languages since India is the world's second most populated and seventh largest country in landmass and a linguistically diverse country. Currently, India has 28 states and 8 Union Territories, where each state and Union Territories has its own language, but none of the languages is recognised as the national language of the country. Only, English and Hindi are used as official languages according to the Constitution of India Part XVII Chapter 1 Article 343\footnote{https://www.mea.gov.in/Images/pdf1/Part17.pdf}. Currently, the Eighth Schedule of the Constitution consists of 22 languages. Table \ref{tab: language} describes the recognised 22 languages according to the Eighth Schedule of the Constitution of India, as of 1 December 2007.
\vspace{-0.5em}
\begin{table}[h!]
\small
\caption{List of languages as per the Eighth Schedule of the Constitution of India, as of 1 December 2007 with their language family \& states spoken in.}
\centering
\label{tab: language}
\begin{tabular}{cccc}
\hline
\textbf{Sl. No.} & \textbf{Language} & \textbf{Family} & \textbf{Spoken in}                                                                                                                                                                                                                                                                         \\ \hline
1                & Assamese          & Indo-Aryan      & Assam                                                                                                                                                                                                                                                                                      \\ [-0.5em]\hdashline
2                & Bengali           & Indo-Aryan      & Assam, Jharkhand, Tripura,  West Bengal                                                                                                                                                                                                                                                    \\[-0.5em] \hdashline
3                & Bodo              & Sino-Tibetan    & Assam                                                                                                                                                                                                                                                                                      \\[-0.5em] \hdashline
4                & Dogri             & Indo-Aryan      & Jammu \& Kashmir                                                                                                                                                                                                                                                                           \\[-0.5em] \hdashline
5                & Gujarati          & Indo-Aryan      & Gujrat, Dadra \& Nagar Haveli \& Daman \& Diu                                                                                                                                                                                                                                              \\[-0.5em] \hdashline
6                & Hindi             & Indo-Aryan      & \begin{tabular}[c]{@{}c@{}}Andaman \& Nicobar Islands, Bihar,  Chhattisgarh, \\[-0.5em] Dadra \& Nagar Haveli \& Daman \& Diu, Delhi, Haryana, \\[-0.5em] Himachal Pradesh,  Jammu \& Kashmir, Jharkhand, \\[-0.5em] Ladakh, Madhya Pradesh, Mizoram, Rajasthan, \\[-0.5em]Uttar Pradesh, Uttarakhand\end{tabular} \\[-0.5em] \hdashline
7                & Kannada           & Dravidian       & Karnataka                                                                                                                                                                                                                                                                                  \\ [-0.5em]\hdashline
8                & Kashmiri          & Indo-Aryan      & Jammu \& Kashmir                                                                                                                                                                                                                                                                           \\[-0.5em] \hdashline
9                & Konkani           & Indo-Aryan      & Dadra \& Nagar Haveli \& Daman \& Diu, Goa                                                                                                                                                                                                                                                 \\[-0.5em] \hdashline
10               & Maithili          & Indo-Aryan      & Jharkhand                                                                                                                                                                                                                                                                                  \\[-0.5em] \hdashline
11               & Malayalam         & Dravidian       & Kerala, Lakshadweep, Puducherry                                                                                                                                                                                                                                                            \\[-0.5em] \hdashline
12               & Manipuri            & Sino-Tibetan    & Manipur                                                                                                                                                                                                                                                                                    \\[-0.5em] \hdashline
13               & Marathi           & Indo-Aryan      & \begin{tabular}[c]{@{}c@{}}Dadra \& Nagar Haveli \& Daman \& Diu,  Goa, \\ [-0.5em]Maharashtra\end{tabular}                                                                                                                                                                                        \\[-0.5em] \hdashline

14               & Nepali            & Indo-Aryan      & Sikkim, West Bengal                                                                                                                                                                                                                                                                        \\[-0.5em] \hdashline
15               & Odia              & Indo-Aryan      & Jharkhand, Odisha                                                                                                                                                                                                                                                                          \\[-0.5em] \hdashline
16               & Punjabi           & Indo-Aryan      & Delhi, Haryana, Punjab                                                                                                                                                                                                                                                                     \\[-0.5em] \hdashline
17               & Sanskrit          & Indo-Aryan      & Himachal Pradesh                                                                                                                                                                                                                                                                           \\[-0.5em] \hdashline
18               & Santali           & Austroasiatic   & Jharkhand                                                                                                                                                                                                                                                                                  \\[-0.5em] \hdashline
19               & Sindhi            & Indo-Aryan      & Rajasthan                                                                                                                                                                                                                                                                                  \\ [-0.5em]\hdashline
20               & Tamil             & Dravidian       & Tamil Nadu                                                                                                                                                                                                                                                                                 \\ [-0.5em]\hdashline
21               & Telugu            & Dravidian       & Andhra Pradesh, Puducherry, Telangana                                                                                                                                                                                                                                                      \\ [-0.5em]\hdashline
22               & Urdu              & Indo-Aryan      & \begin{tabular}[c]{@{}c@{}}Bihar, Delhi, Jammu \& Kashmir, \\[-0.5em]  Jharkhand, Telangana, Uttar Pradesh\end{tabular}                                                                                                                                                                            \\ \hline
\end{tabular}
\end{table}

Most of the Indian languages originated from the Indo-Aryan and Dravidian language families.
It can be seen from Table \ref{tab: language} that different languages are spoken in different states, however, languages do not obey geographical boundaries. Therefore, many of these languages, particularly in the neighbouring regions, have multiple dialects which are amalgamations of two or more languages.

Such enormous linguistic diversity makes it difficult for citizens to communicate in different parts of the country. Bilingualism and multilingualism are the norms in India. In this context, a LID system becomes a crucial component for any speech-based smart assistant. The biggest challenge and hence an area of active innovation for the Indian language is the reality that most of these languages are under-resourced.

Every spoken language has its underlying lexical, speaker, channel, environment, and other variations. The likely differences among various spoken languages are in their phoneme inventories, frequency of occurrence of the phonemes, acoustics, the span of the sound units in different languages, and intonation patterns at higher levels. The overlap between the phoneme set of two or more familial languages makes it a challenge for recognition. The low-resource status of these languages makes the training of machine learning models doubly difficult. The idea behind our methodology is interesting on account of the aforementioned limitations. Our methodology involves forecasting the accurate spoken language, irrespective of the limitations mentioned earlier.

CNN has been heavily utilized by Natural Language Processing (NLP) researchers from the very beginning due to their efficient use of local features. While Recurrent Neural Networks (RNNs) have been shown to be effective in a variety of NLP tasks in the past, recent work with Attention-based methods have outperformed all previous models and architectures  because of their ability to capture global interactions. \cite{Yamada2020} were able to achieve better results than BERT \citep{Devlin2019}, SpanBERT \citep{Joshi2020}, XLNet \citep{Yang2019}, and ALBERT \citep{Zhenzhong2020} using their Attention-based methods in the Question-Answering domain. Researchers \citep{Takase2021, Gu2019, Chen2020} have employed Attention-based methods to achieve state-of-the-art (SOTA) performance in Machine Translation. Transformers \citep{Vaswani2017}, which utilize a self-attention mechanism, have found extensive application in almost all fields of NLP such as language modelling, text classification, topic modelling, emotion classification, sentiment analysis, etc., and produced SOTA performance.

In this work, we present LID for Indian languages using a combination of CNN, RNN, and Attention-based methods. Our LID methods cover 13 Indian languages\footnote{The study was limited to the number of Indian languages for which datasets were available}. Additionally, our method is language agnostic. The main contributions of this work can be summarized as follows:
\begin{itemize}
    \vspace{-1.0em}
    \item We carried out exhaustive experiments using CNN, CRNN, and attention-based CRNN for the LID task on 13 Indian languages and achieved state-of-the-art results.
    \item The model exhibits exceptional performance in languages that are part of the same language family, as well as in diverse language sets under both normal and noisy conditions.
    \item We empirically proved that CRNN framework achieves better or similar results compared to CRNN with Attention framework although CRNN without Attention requires less computational overhead.
\end{itemize} 

\section{Related Works}
\label{sec:Rel_Works}
Extraction of language-dependent features for example prosody and phonemes was widely used to classify spoken languages \citep{Zissman1996, Martinez2011, Ferrer2010}. Following the success of speaker verification systems, identity vectors (i-vectors) have also been used as features in various classification frameworks. Use of i-vectors requires significant domain knowledge \citep{Dehak2011b, Martinez2011}. In recent trends, researchers rely on neural networks for feature extraction and classification \citep{Lopez2014, Ganapathy2014}. Researcher \citet{Revay2019} used the ResNet50 \citep{He2016} framework for classifying languages by generating the log-Mel spectra for each raw audio. The framework uses a cyclic learning rate where the learning rate increases and then decreases linearly. The maximum learning rate for a cycle is set by finding the optimal learning rate using fastai \citep{Howard2020}.

Researcher \citet{Gazeau2018} established the use of a Neural Network, Support Vector Machine, and Hidden Markov Model (HMM) to identify different languages. Hidden Markov models convert speech into a sequence of vectors and are used to capture temporal features in speech. Established LID systems \citep{Dehak2011a, Martinez2011, Plchot2016, Zazo2016} are based on identity vector (i-vectors) representations for language processing tasks. In \citet{Dehak2011a}, i-vectors are used as data representations for a speaker verification task and fed to the classifier as the input. \citet{Dehak2011a} applied Support Vector Machines (SVM) with cosine kernels as the classifier, while \citet{Martinez2011} used logistic regression for the actual classification task. Recent years have found the use of feature extraction with neural networks, particularly with Long Short Term Memory (LSTM) \citep{Zazo2016, Gelly2016, Diez2015}. These neural networks produce better accuracy while being simpler in design compared to the conventional LID methods \citep{Dehak2011a, Martinez2011, Plchot2016}. Recent trends in developing LID systems are mainly focused on different forms of LSTMs with DNNs. \citet{Plchot2016} used a 3-layered CNN where i-vectors were the input layer and softmax activation function was the output layer. \citet{Zazo2016} used MFCC with Shifted Delta Coefficient features as information to a unidirectional layer that is directly connected to a softmax classifier. \citet{Gelly2016} used audio transformed to Perceptual Linear Prediction (PLP) coefficients and their $1^{st}$ and $2^{nd}$ order derivatives as information for a Bidirectional LSTM in forward and backward directions.  The forward and backward sequences generated from the Bidirectional LSTM were joined together and used to classify the language of the input samples. \citet{Diez2015} used CNNs for their LID system. They transformed the input data into an image containing MFCCs with Shifted Delta Coefficient features. The image represents the time domain for the x-axis and frequency bins for the y-axis. 

\citet{Diez2015} used CNN as the feature extractor for the identity vectors. They achieved better performance when combining both the CNN features and identity vectors. \citet{Revay2019} used ResNet \citep{He2016} framework for language classification by generating spectrograms of each audio. Cyclic Learning \citep{Smith2018} was used where the learning rate increases and decreases linearly. \citet{Venkatesan2018} utilised MFCCs to infer aspects of speech signals from Kannada, Hindi, Tamil, and Telugu. They obtained an accuracy of 76\% and 73\% using Support Vector Machines and Decision Tree classifiers, respectively, on 5 hours of training data. \citet{Mukherjee2019} used CNNs for language identification in German, Spanish, and English. They used Filter Banks to extract features from frequency domain representations of the signal. \citet{Aarti2017} experimented with several auditory features in order to determine the optimal feature set for a classifier to detect Indian Spoken Language. \citet{Sisodia2020} evaluated Ensemble Learning models for classifying spoken languages such as German, Dutch, English, French, and Portuguese. Bagging, Adaboosting, random forests, gradient boosting, and additional trees were used in their ensemble learning models.

\citet{Heracleous2018} presented a comparative study of Deep Neural Networks (DNN) and CNNs for Spoken LID, with Support Vector Machines (SVM) as the baseline. They also presented the performance of the fusion of the mentioned methods. The NIST 2015 i-vector Machine Learning Challenge dataset was used to assess the system's performance with the goal of detecting 50 in-set languages. \citet{Bartz2017} tackled the problem of Language Identification in the image domain rather than the typical acoustic domain. A hybrid CRNN is employed for this, which acts on spectrogram images of the provided audio clips. \citet{Draghichi2020} tried to solve the task of Language Identification while using Mel-spectrogram images as input features. This strategy was employed in CNNs and CRNN in terms of performance. This work is characterized by a modified training strategy that provides equal class distribution and efficient memory utilisation. \citet{Ganapathy2014} reported how they used bottleneck features from a CNN for the LID task. Bottleneck features were used in conjunction with conventional acoustic features, and performance was evaluated. Experiments revealed that when a system with bottleneck features is compared to a system without them, average relative improvements of up to 25\% are achieved. \citet{Zazo2016} proposed an open-source, end-to-end, LSTM-RNN system that outperforms a more recent reference i-vector system by up to 26\% when both are tested on a subset of the NIST Language Recognition Evaluation with 8 target languages.

Our research differs from the previous works on LID in the following aspects:

\begin{itemize}
    \vspace{-1em}
    \item Comparison of performance of CNN, CRNN, as well as CRNN with Attention.
    \item Extensive experiments with our proposed model show its applicability both for close language as well as noisy speech scenarios.
\end{itemize}

\section{Model Framework}
\label{sec:Model_Architecture}
Our proposed framework consists of three models.

\begin{itemize}
    \vspace{-1.0em}
    \item CNN-based framework
    \item CRNN-based framework
    \item CRNN with Attention-based framework
    \vspace{-0.5em}
\end{itemize}

We made use of the capacity of CNNs to capture spatial information to identify languages from audio samples. In a CNN-based framework, our network uses four convolution layers, where each layer is followed by the ReLU \citep{Nair2010} activation function and max pooling with a stride of 3 and a pool size of 3. The kernel sizes and the number of filters for each convolution layer are (3, 512), (3, 512), (3, 256), and (3, 128), respectively.

\begin{figure}[h!]
\centering
\includegraphics[width=\textwidth,keepaspectratio]{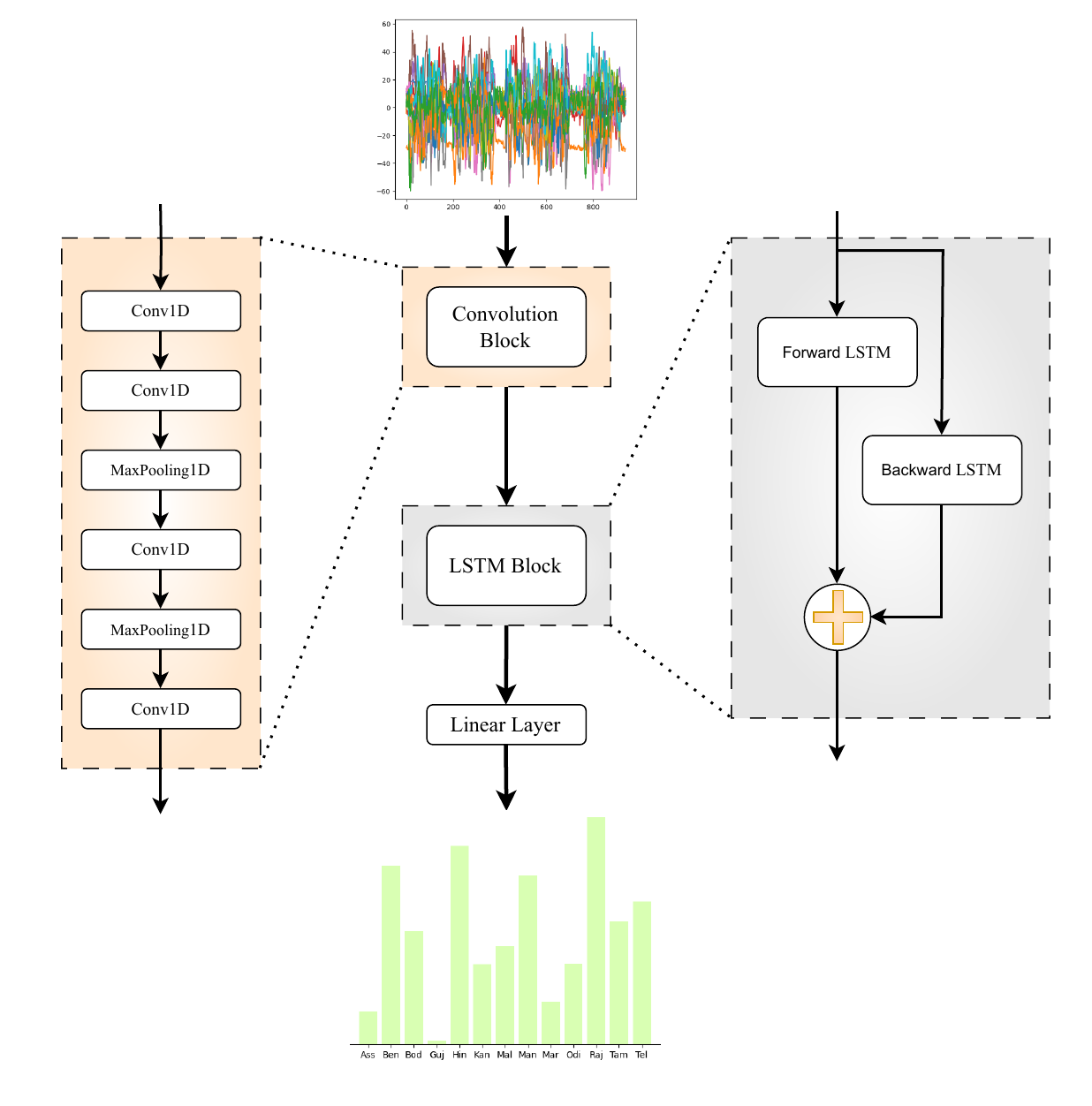}
\caption{The figure presents our CRNN framework consisting of a Convolution Block and LSTM Block denoted in different blocks. The convolution block extracts feature from the input audio. The output of the final convolution layer is provided to the Bi-Directional LSTM network as the input which is further connected to a Linear Layer with softmax classifier.}
\label{fig:framework_model}
\end{figure}

Figure \ref{fig:framework_model} provides a schematic overview of the framework. The CRNN framework passes the output of the Convolutional Module to a Bi-Directional LSTM consisting of a single LSTM with 256 output units. The LSTM's activation function is $tanh$, and its recurrent activation is $sigmoid$. The Attention Mechanism used in our framework is based on Hierarchical Attention Networks \citep{Yang2016}. In the Attention Mechanism, contexts of features are summarized with a bidirectional LSTM by going forward and backwards.

\begin{subequations}
\begin{equation}
\overrightarrow{a_{n}} = \overrightarrow{LSTM}(a_n), n \in [ 1, L ]
\end{equation}
\begin{equation}
\overleftarrow{a_{n}} = \overleftarrow{LSTM}(a_n), n \in [ L, 1 ]
\end{equation}
\begin{equation}
\label{eq: forward-backward}
a_{i} = [  \overrightarrow{a_{n}}, \overleftarrow{a_{n}} ]
\end{equation}
\label{eq: LSTMperceptron}
\end{subequations}

In equation \ref{eq: LSTMperceptron}, $L$ is the number of audio specimens. $a_{n}$ is the input sequence for the LSTM network. $\overrightarrow{a_{n}}$ and $\overleftarrow{a_{n}}$ provide the learned vectors from LSTM forward direction and backward direction, respectively. The vector, $a_{i}$, builds the base for the attention mechanism. The goal of the attention mechanism is to learn the model through training with randomly initialized weights and biases. The layer also ensures with the $tanh$ function that the network does not stall. The function keeps the input values between -1 and 1 and maps zeros to near-zero values. The layer with $tanh$ function is again multiplied by trainable context vector $u_{i}$. The trainable context vector refers to a vector learned during the training process and used as a fixed-length representation of the entire input document. In our framework, the attention mechanism is used to compute a weighted sum of the sequences for each speech utterance, where the weights are learned based on the relevance of each sequence to the speech utterances. This produces a fixed-length vector for each utterance that captures the most salient information in the sequences. The context weight vector $u_{i}$ is randomly initiated and jointly learned during the training process. Improved vectors are represented by $a_{i}^{'}$ as shown in equation \ref{eq: annotationImp}.

\begin{equation}
\label{eq: annotationImp}
    a_{i}^{'} = tanh(a_{i}\cdot w_{i}+b_{i})\cdot u_{i}
\end{equation}

Context vectors are finally calculated by providing a weight to each $W_{i}$ by dividing the exponential values of the previously generated vectors with the summation of all exponential values of previously generated vectors as shown in equation \ref{eq: Attention}. To avoid division by zero, an epsilon is added to the denominator.
\begin{equation}
\label{eq: Attention}
    W_{i} = \frac{exp(a^{'}_{i})}{\sum_{i}exp(a_{i}^{'}) + \epsilon}
\end{equation}

The sum of these importance weights concatenated with the previously calculated context vectors is fed to a linear layer with 13 output units serving as a classifier for the 13 languages.

\begin{figure}[h!]
\centering
\includegraphics[width=\textwidth,keepaspectratio]{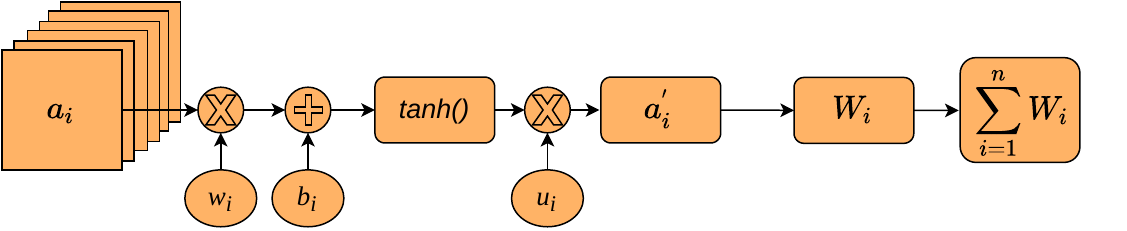}
\caption{Schematic diagram of the Attention Module.}
\label{fig:attention_model}
\end{figure}

Figure \ref{fig:attention_model} presents the schematic diagram of the Attention Module where $a_{i}$ is the input to the module and output of the Bi-Directional LSTM layers.

\section{Experiments}
\label{sec:Experiments}
\subsection{Feature Extraction}
\label{subsec:Feature_Extraction}
For feature extraction of spoken utterances, we used MFCCs. For calculating MFCCs we used $pre\_emphasis$, frame size represented as $f\_size$, frame stride represented as $f\_stride$,  N-point Fast Fourier transform represented as $NFFT$, low-frequency mel represented as $lf$, the number of filters represented as $nfilt$, the number of cepstral coefficients represented as $ncoef$ and cepstral lifter represented $lifter$ of values 0.97, 0.025 (25ms), 0.015 (15ms overlapping), 512, 0, 40, 13, and 22, respectively. We used a frame size of 25 ms as typically frame sizes in the speech processing domain use 20ms to 40ms with 50\% (in our case 15ms) overlapping between consecutive frames.
\begin{equation}
\label{eq: hf}
    hf = 2595 \times \log_{10}(1 + \frac{0.5 \times sr}{700})
\end{equation}
We used low-frequency mel (lf) as 0 and high-frequency mel (hf) is calculated using the equation \ref{eq: hf}. lf and hf are used to generate the non-linear human ear perception of sound, by being more discriminative at lower frequencies and less discriminative at higher frequencies.
\begin{equation}
\label{eq: es}
    emphasized\_signal = [sig[0], sig[1:] - pre\_emphasis * sig[:-1]]
\end{equation}
As shown in equation \ref{eq: es} emphasized signal is calculated by using a pre-emphasis filter applied on the signal ($sig$) using the first-order filter. The number of frames is calculated by taking the ceiling value of the division of the absolute value of the difference between signal length ($sig\_len$) and product of filter size ($f\_size$) and sample rate (sr) with the product of frame stride ($f\_stride$) and sample rate (sr) as shown in equation \ref{eq: nframes}. Signal length is the length of $emphasized\_signal$ calculated in equation \ref{eq: es}.
\begin{equation}
\label{eq: nframes}
    n\_frames = \lceil \frac{\lvert sig\_len - (f\_size \times sr) \rvert}{(f\_stride \times sr)} \rceil
\end{equation}
Using equation \ref{eq: pad_sig} $pad\_signal$ is generated from concatenation of $emphasized\_signal$ and zero value array of dimension ($pad\_signal\_length - signal\_length$)$\times$1, where, $pad\_signal\_length$ is calculated by $n\_frames\times (f\_stride \times sr) + (f\_size \times sr)$. 
\begin{equation}
\label{eq: pad_sig}
    pad\_signal = [emphasized\_signal, [0]_{((n\_frames\times (f\_stride \times sr) + (f\_size \times sr)) - sig\_len)\times1}]
\end{equation}

Frames are calculated as shown in equation \ref{eq: frames} from the $pad\_signal$ elements where elements are the addition of an array of positive natural numbers from 0 to $f\_size\times sr$ repeated $n\_frames$ and the transpose of the array of size of $num\_frames$ where each element is the difference of $(f\_stride\times sr)$.

\begin{equation}
\label{eq: frames}
\begin{gathered}
    frames = pad\_signal[(\{ x \in Z^{+} : 0 < x < (f\_size \times sr) \}_{n})_{n=0}^{(n\_frames, 1)} + \\
    ((\{r : r =  (f\_stride \times sr)\times(i-1), i \in \{0,\ldots, n\_frames \times (f\_stride \times sr)\} \}_{n})_{n=0}^{((f\_size \times sr), 1)})^\mathrm{T}]
\end{gathered}
\end{equation}

Power frames shown in equation \ref{eq: pf} are calculated as the square of the absolute value of the Discrete Fourier Transform (DFT) of the product of hamming window and frames of each element with NFFT.
\begin{equation}
\label{eq: pf}
    pf = \frac{\lvert DFT((frames \times (0.54 - (\sum_{N = 0}^{(f\_size \times sr)-1}0.46\times\cos{\frac{2\pi N}{(f\_size \times sr)-1}}))), NFFT)\rvert^{2}}{NFFT} 
\end{equation}

\begin{equation}
\label{eq: mel_points}
    mel\_points = \{r: r = lf + \frac{hf-lf}{(nfilt+2)-1} \times i, i \in \{lf, \ldots, hf\} \}
\end{equation}
Mel points are the array where elements are calculated as shown in the equation \ref{eq: mel_points}, where i is the values belonging from lf to hf.
\begin{equation}
\label{eq: bins}
    bins = \lfloor\frac{(NFFT + 1)\times (700 \times (10^{\frac{mel\_points}{2595}} - 1)) }{sample\_rate}\rfloor
\end{equation}
From equation \ref{eq: bins}, bins are calculated where the floor value of the elements are taken which is the product of hertz points and $NFFT + 1$ divided by the sample rate. Hertz points are calculated by multiplying 700 by subtraction of 1 from 10 power of $\frac{mel\_points}{2595}$.
\begin{equation}
\label{eq: fbank}
    fbank_{m}(k) = \begin{cases}
        0 & k < bins(m-1)\\
        \frac{k - bins(m-1)}{bins(m) - bins(m-1)} & bins(m-1) \leq k \leq bins(m)\\
        \frac{bins(m+1) - k}{bins(m+1) -bins(m)} & bins(m) \leq k \leq bins(m+1)\\
        0 & k > bins(m+1)
    \end{cases}
\end{equation}
Bins calculated from equation \ref{eq: bins} are used to calculate filter banks shown in equation \ref{eq: fbank}. Each filter in the filter bank is triangular, with a response of 1 at the central frequency and a linear drop to 0 till it meets the central frequencies of the two adjacent filters, where the response is 0. \par
Finally, mfcc is calculated shown in equation \ref{eq: mfcc} by decorrelating the filter bank coefficients using Discrete Cosine Transform (DCT) to get a compressed representation of the filter banks. Sinusoidal liftering is applied to the mfcc to de-emphasize higher mfccs which improves classification in noisy signals.
\begin{equation}
\label{eq: mfcc}
    mfcc = DCT(20\log_{10}(pf\cdot fbank^\mathrm{T})) \times \left[1 + \frac{lifter}{2}\sin{\frac{\{ \pi \odot n : n \in Z^{+}, n \leq ncoef \}}{lifter}}\right]
\end{equation}

MFCCs features of shape $(1000, 13)$ generated from equation \ref{eq: mfcc} is provided as input to the neural network which expects the same dimension followed by convolution layers as mentioned in section \ref{sec:Model_Architecture}. Raw speech signal cannot be provided input to the framework as it contains lots of noise data therefore extracting features from the speech signal and using it as input to the model will produce better performance than directly considering raw speech signal as input. Our motivation to use MFCC features as the feature count is small enough to force us to learn the information of the sample. Parameters are related to the amplitude of frequencies and provide us with frequency channels to analyze the speech specimen.

\subsection{Data}
\label{subsec:Data}
\subsubsection{Benchmark Data}
\label{subsubsec:Benchmark_data}
The Indian language (\textit{IL}) dataset was acquired from the Indian Institute of Technology, Madras\footnote{https://www.iitm.ac.in/donlab/tts/database.php}. The dataset includes 13 widely used Indian languages. Table \ref{tab:  dataset} presents the statistics of this dataset which we used for our experiments.

\begin{table}[h!]
\small
\centering
\caption{Statistics of the Indian Language (\textit{IN}) Dataset}
\label{tab:  dataset}

\begin{tabular}{cccccc}
\hline
\textbf{Language} & \textbf{Label} & \textbf{Gender}  & \textbf{Samples} & \textbf{Total Sample{s}} & \textbf{\begin{tabular}[c]{@{}c@{}}Average Duration \\[-0.5em] (in seconds) \end{tabular}}\\
\hline
\multirow{2}{*}{Assamese}   & \multirow{2}{*}{as} & F  & 8,713 & \multirow{2}{*}{17,654} & \multirow{2}{*}{5.587} \\ \cdashline{3-4}
                            &                     & M  & 8,941 & &                         \\[-0.5em]\hdashline
\multirow{2}{*}{Bengali}    & \multirow{2}{*}{bn} & F  & 3,253 & \multirow{2}{*}{9,440}  & \multirow{2}{*}{5.743} \\ \cdashline{3-4}
                            &                     & M  & 6,187 & &                        \\[-0.5em]\hdashline
    Bodo                        & bd                  & F  & 571  & 571 & 25.219                   \\[-0.5em]\hdashline
\multirow{2}{*}{Gujarati}    & \multirow{2}{*}{gu} & F  & 2,396 & \multirow{2}{*}{5,684} & \multirow{2}{*}{13.459} \\ \cdashline{3-4}
                            &                     & M & 3,288 & &                         \\[-0.5em] \hdashline
\multirow{2}{*}{Hindi}      & \multirow{2}{*}{hi} & F & 2,318 & \multirow{2}{*}{4,636}  & \multirow{2}{*}{8.029}\\ \cdashline{3-4}
                            &                     & M & 2,318 &                      &  \\ [-0.5em]\hdashline
\multirow{2}{*}{Kannada}    & \multirow{2}{*}{kn} & F & 1,289 & \multirow{2}{*}{2,578}  & \multirow{2}{*}{10.264}\\ \cdashline{3-4}
                            &                     & M & 1,289 &                &        \\[-0.5em] \hdashline
\multirow{2}{*}{Malayalam}  & \multirow{2}{*}{ml} & F & 5,650 & \multirow{2}{*}{11,300} & \multirow{2}{*}{5.699} \\ \cdashline{3-4}
                            &                     & M & 5,650 &                &        \\ [-0.5em]\hdashline
\multirow{2}{*}{Manipuri}   & \multirow{2}{*}{mn} & F & 9,487 & \multirow{2}{*}{17,917} & \multirow{2}{*}{4.169} \\ \cdashline{3-4}
                            &                     & M & 8,430 &                &        \\ [-0.5em]\hdashline
Marathi                     & mr                  & F & 2,448 & 2,448               &    7.059 \\ [-0.5em]\hdashline
\multirow{2}{*}{Odia}       & \multirow{2}{*}{or} & F & 3,578 & \multirow{2}{*}{7,151} & \multirow{2}{*}{4.4} \\ \cdashline{3-4}
                            &                     & M & 3,573 &               &         \\ [-0.5em]\hdashline
\multirow{2}{*}{Rajasthani} & \multirow{2}{*}{rj} & F & 4,346 & \multirow{2}{*}{9,125}  & \multirow{2}{*}{7.914}\\ \cdashline{3-4}
                            &                     & M & 4,779 &                &        \\ [-0.5em]\hdashline
\multirow{2}{*}{Tamil}      & \multirow{2}{*}{ta} & F & 3,243 & \multirow{2}{*}{6,960}  & \multirow{2}{*}{10.516} \\ \cdashline{3-4}
                            &                     & M & 3,717 &                       & \\ [-0.5em]\hdashline
\multirow{2}{*}{Telugu}     & \multirow{2}{*}{te} & F & 4,043 & \multirow{2}{*}{6,524}  & \multirow{2}{*}{15.395}\\ \cdashline{3-4}
                            &                     & M & 2,481 &                &        \\
\hline
\end{tabular}
\end{table}

\subsubsection{Experimental Data}
\label{subsubsec:experimental_data}
In the past two decades, the development of LID methods has been largely fostered through NIST Language Evaluations (LREs). As a result, the most popular benchmark for evaluating new LID models and methods is the NIST LRE evaluation dataset \citep{Sadjadi2018}. The NIST LREs dataset mostly contains narrow-band telephone speech. Datasets are typically distributed by the Linguistic Data Consortium (LDC) and cost thousands of dollars. For example, the standard Kaldi \citep{Povey2011} recipe for LRE072 relies on 18 LDC SLR datasets that cost \$15000 (approx) to LDC non-members. This makes it difficult for new research groups to enter the academic field of LID. Furthermore, the NIST LRE evaluations focus mostly on telephone speech.
\begin{table}[h!]
\small
\caption{Statistics of the EU Dataset}
\centering
\label{tab: eu_dataset}
\begin{tabular}{cccc}
\hline
\textbf{Language} & \textbf{Label} & \textbf{Total Samples} & \textbf{\begin{tabular}[c]{@{}c@{}}Average Duration\\[-0.5em] (in seconds)\end{tabular}} \\ \hline
English           & en             & 43,269                 & 684.264                                                                          \\ \hdashline
French            & fr             & 67,689                 & 492.219                                                                          \\ \hdashline
German            & de             & 48,454                 & 1,152.916                                                                        \\ \hdashline
Spanish           & es             & 57,869                 & 798.169                                                                          \\ \hline
\end{tabular}
\end{table}

As the NIST LRE dataset is not freely available we used the EU Dataset \citep{Bartz2017} which is open source. The (\textit{EU}) dataset contains YouTube News data for 4 major European languages -- English (\textit{en}), French (\textit{fr}), German (\textit{de}) and Spanish (\textit{es}). Statistics of the dataset are given in Table \ref{tab: eu_dataset}.

\subsection{Environment}
\label{subsec:environment}
We implemented our framework using Tensorflow \citep{Abadi2016} backend. We split the Indian language dataset into training, validation, and testing set, containing 80\%, 10\%, and 10\% of the data, respectively, for each language and gender.

For regularization, we apply dropout \citep{Srivastava2014} after the Max-Pooling layer and Bi-Directional LSTM layer. We use the rate of 0.1. A $l_{2}$ regularization with $10^{-6}$ weight is also added to all the trainable weights in the network. We train the model with Adam \citep{Kingma2014} optimizer with $\beta_{1} = 0.9$, $\beta_{2} = 0.98$, and $\epsilon = 10^{-9}$ and learning rate schedule \citep{Vaswani2017}, with 4k warm-up steps and peak learning rate of $0.05/\sqrt{d}$ where d is 128. A batch size of 64 with ``Sparse Categorical Crossentropy'' as the loss function was used.

\subsection{Result on Indian Language Dataset}
\label{subsec:results_indian}

The proposed framework was assessed against  \citet{Kulkarni2022} using identical datasets. Both CRNN and CRNN with Attention exhibited superior performance compared to the results reported by \citet{Kulkarni2022}, as shown in Table \ref{tab: eval_13}. They used 6 Linear layers where units are 256, 256, 128, 64, 32, and 13, respectively in the CNN framework, whereas the DNN framework uses 3 LSTM layers having units 256, 256, and 128, respectively followed by dropout layer followed by 3 Time Distributed layer followed by a Linear layer of 13 as units.
\begin{table}[h!]
\small
\caption{Comparative evaluation results (in terms of Accuracy) of our model and the model of \citet{Kulkarni2022} on the Indian Language dataset}
\centering
\label{tab: eval_13}
\begin{tabular}{cc}
\hline
     & \textbf{Accuracy} \\
\hline
DNN \citep{Kulkarni2022} & 0.9834\\[-0.5em]\hdashline
RNN \citep{Kulkarni2022} & 0.9843\\[-0.5em]\hdashline
CNN & 0.983 \\[-0.5em]\hdashline
CRNN & \textbf{0.987} \\[-0.5em]\hdashline
CRNN with Attention & \textbf{0.987}\\
\hline
\end{tabular}
\end{table}

We evaluated system performance using the following evaluation metrics -- Recall (TPR), Precision (PPV), f1-score, and Accuracy. Since one of our major objectives was to measure the accessibility of the network to new languages, we introduced Data Balancing of training data for each class, as the number of samples available for each class may vary drastically. This is the case for the Indian Language Dataset as shown in Table \ref{tab: dataset} in which Kannada, Marathi and particularly Bodo have a limited amount of data compared to the rest of the languages. To alleviate this data imbalance problem, we used class weight balancing as a dynamic method using scikit-learn \citep{Pedregosa2011}.

\begin{table}[h!]
\tiny 
\caption{Experimental Results for Indian Languages}
\centering
\label{tab: res_lang_class_weight}
\begin{tabular}{c|c|c|c|c|c|c|c|c|c|c|c|c}
\hline
\multirow{2}{*}{\textbf{Language}} & \multicolumn{4}{c|}{\textbf{CRNN with Attention}} & \multicolumn{4}{c|}{\textbf{CRNN}} & \multicolumn{4}{c}{\textbf{CNN}} \\ \cline{2-13}
    & \textbf{PPV} & \textbf{TPR} & \textbf{\begin{tabular}[c]{@{}c@{}}f1\\[-0.5em] Score\end{tabular}} & \textbf{Accuracy} & \textbf{PPV} & \textbf{TPR} & \textbf{\begin{tabular}[c]{@{}c@{}}f1\\[-0.5em] Score\end{tabular}} & \textbf{Accuracy} & \textbf{PPV} & \textbf{TPR} & \textbf{\begin{tabular}[c]{@{}c@{}}f1\\[-0.5em] Score\end{tabular}} & \textbf{Accuracy}\\
\hline
\textbf{as} & 0.989 & 0.998 & 0.993 & \multirow{13}{*}{0.987} & 0.995 & 0.989 & 0.992 & \multirow{13}{*}{0.987} & 0.991 & 0.988 & 0.989 & \multirow{13}{*}{0.983} \\ \cline{1-4} \cline{6-8} \cline{10-12}
\textbf{bn} & 1 & 0.9 & 0.948 &  & 1 & 0.904 & 0.949 &  & 1 & 0.888 & 0.941 &  \\ \cline{1-4} \cline{6-8} \cline{10-12}
\textbf{bd} & 0.966 & 1 & 0.983 &  & 0.966 & 1 & 0.983 &  & 0.966 & 1 & 0.983 &  \\ \cline{1-4} \cline{6-8} \cline{10-12}
\textbf{gu} & 0.997 & 0.997 & 0.997 &  & 0.951 & 0.998 & 0.974 &  & 0.996 & 0.991 & 0.994 &  \\ \cline{1-4} \cline{6-8} \cline{10-12}
\textbf{hi} & 0.987 & 0.991 & 0.989 &  & 0.991 & 0.991 & 0.991 &  & 0.991 & 0.974 & 0.983 &  \\ \cline{1-4} \cline{6-8} \cline{10-12}
\textbf{kn} & 0.977 & 0.996 & 0.987 &  & 0.996 & 0.996 & 0.996 &  & 0.973 & 0.992 & 0.983 &  \\ \cline{1-4} \cline{6-8} \cline{10-12}
\textbf{ml} & 0.996 & 0.988 & 0.992 &  & 0.997 & 0.99 & 0.994 &  & 0.99 & 0.993 & 0.991 &  \\ \cline{1-4} \cline{6-8} \cline{10-12}
\textbf{mn} & 0.987 & 0.999 & 0.993 &  & 0.973 & 0.999 & 0.986 &  & 0.972 & 0.998 & 0.985 &  \\ \cline{1-4} \cline{6-8} \cline{10-12}
\textbf{mr} & 1 & 1 & 1 &  & 1 & 1 & 1 &  & 1 & 0.996 & 0.998 &  \\ \cline{1-4} \cline{6-8} \cline{10-12}
\textbf{or} & 1 & 1 & 1 &  & 1 & 0.999 & 0.999 &  & 0.986 & 0.999 & 0.992 &  \\ \cline{1-4} \cline{6-8} \cline{10-12}
\textbf{rj} & 0.999 & 0.993 & 0.996 &  & 0.995 & 1 & 0.997 &  & 0.986 & 0.996 & 0.991 &  \\ \cline{1-4} \cline{6-8} \cline{10-12}
\textbf{ta} & 0.929 & 0.991 & 0.959 &  & 0.975 & 0.997 & 0.986 &  & 0.946 & 0.989 & 0.967 &  \\ \cline{1-4} \cline{6-8} \cline{10-12}
\textbf{te} & 0.979 & 0.998 & 0.989 &  & 0.982 & 1 & 0.991 &  & 0.975 & 0.998 & 0.986 &  \\
\hline
\end{tabular}
\end{table}

PPV, TPR, f1-score, and Accuracy scores are reported in Table \ref{tab: res_lang_class_weight} for the three frameworks - CNN, CRNN, and CRNN with Attention. From Table \ref{tab: res_lang_class_weight} it is clearly visible that both CRNN framework and CRNN with Attention provide competitive results of 0.987 accuracy. Table \ref{tab: 13_category_full_Atten}, Table \ref{tab: 13_category_full_CRNN}, and Table \ref{tab: 13_category_full_cnn} shows the confusion matrix for CNN, CRNN, and CRNN with Attention.

\begin{table}[h!]
\tiny
\caption{Confusion matrix for CRNN with Attention framework}
\centering
\label{tab: 13_category_full_Atten}
\begin{tabular}{c|c|c|c|c|c|c|c|c|c|c|c|c|c|c|c|c|c}
\hline
\multicolumn{2}{c|}{\multirow{2}{*}{}} & \multicolumn{13}{c|}{\textbf{Predicted}} & \multirow{2}{*}{\textbf{PPV}} & \multirow{2}{*}{\textbf{TPR}} & \multirow{2}{*}{\textbf{\begin{tabular}[c]{@{}c@{}}f1\\[-0.5em] Score\end{tabular}}} \\ \cline{3-15}
\multicolumn{2}{c|}{} & \textbf{as} & \textbf{bn} & \textbf{bd} & \textbf{gu} & \textbf{hi} & \textbf{kn} & \textbf{ml} & \textbf{mn} & \textbf{mr} & \textbf{or} & \textbf{rj} & \textbf{ta} & \textbf{te} &  &  &  \\
\hline
\parbox[t]{2mm}{\multirow{13}{*}{\rotatebox[origin=c]{90}{\textbf{Actual}}}} & \textbf{as} & \textbf{1762} & 0 & 0 & 0 & 0 & 0 & 0 & 4 & 0 & 0 & 0 & 0 & 0 & 0.989 & 0.998 & 0.993 \\ \cline{2-18} 
    & \textbf{bn} & 10 & \textbf{850} & 0 & 1 & 0 & 0 & 0 & 18 & 0 & 0 & 0 & 53 & 12 & 1 & 0.9 & 0.948 \\ \cline{2-18} 
    & \textbf{bd} & 0 & 0 & \textbf{57} & 0 & 0 & 0 & 0 & 0 & 0 & 0 & 0 & 0 & 0 & 0.966 & 1 & 0.983 \\ \cline{2-18} 
    & \textbf{gu} & 1 & 0 & 0 & \textbf{566} & 0 & 0 & 0 & 0 & 0 & 0 & 0 & 0 & 1 & 0.997 & 0.997 & 0.997 \\ \cline{2-18} 
    & \textbf{hi} & 0 & 0 & 0 & 0 & \textbf{460} & 0 & 4 & 0 & 0 & 0 & 0 & 0 & 0 & 0.987 & 0.991 & 0.989 \\ \cline{2-18} 
    & \textbf{kn} & 0 & 0 & 1 & 0 & 0 & \textbf{257} & 0 & 0 & 0 & 0 & 0 & 0 & 0 & 0.977 & 0.996 & 0.987 \\ \cline{2-18} 
    & \textbf{ml} & 0 & 0 & 1 & 0 & 6 & 5 & \textbf{1116} & 1 & 0 & 0 & 0 & 0 & 1 & 0.996 & 0.988 & 0.992 \\ \cline{2-18} 
    & \textbf{mn} & 0 & 0 & 0 & 1 & 0 & 0 & 0 & \textbf{1790} & 0 & 0 & 0 & 0 & 0 & 0.987 & 0.999 & 0.993 \\ \cline{2-18} 
    & \textbf{mr} & 0 & 0 & 0 & 0 & 0 & 0 & 0 & 0 & \textbf{245} & 0 & 0 & 0 & 0 & 1 & 1 & 1 \\ \cline{2-18} 
    & \textbf{or} & 0 & 0 & 0 & 0 & 0 & 0 & 0 & 0 & 0 & \textbf{716} & 0 & 0 & 0 & 1 & 1 & 1 \\ \cline{2-18} 
    & \textbf{rj} & 4 & 0 & 0 & 0 & 0 & 1 & 1 & 0 & 0 & 0 & \textbf{906} & 0 & 0 & 0.999 & 0.993 & 0.996 \\ \cline{2-18} 
    & \textbf{ta} & 5 & 0 & 0 & 0 & 0 & 0 & 0 & 0 & 0 & 0 & 1 & \textbf{690} & 0 & 0.929 & 0.991 & 0.959 \\ \cline{2-18} 
    & \textbf{te} & 0 & 0 & 0 & 0 & 0 & 0 & 0 & 1 & 0 & 0 & 0 & 0 & \textbf{652} & 0.979 & 0.998 & 0.989 \\
\hline
\end{tabular}
\end{table}

\begin{table}[h!]
\tiny
\caption{Confusion matrix for CRNN}
\centering
\label{tab: 13_category_full_CRNN}
\begin{tabular}{c|c|c|c|c|c|c|c|c|c|c|c|c|c|c|c|c|c}
\hline
\multicolumn{2}{c|}{\multirow{2}{*}{}} & \multicolumn{13}{c|}{\textbf{Predicted}} & \multirow{2}{*}{\textbf{PPV}} & \multirow{2}{*}{\textbf{TPR}} & \multirow{2}{*}{\textbf{\begin{tabular}[c]{@{}c@{}}f1\\[-0.5em] Score\end{tabular}}} \\ \cline{3-15}
\multicolumn{2}{c|}{} & \textbf{as} & \textbf{bn} & \textbf{bd} & \textbf{gu} & \textbf{hi} & \textbf{kn} & \textbf{ml} & \textbf{mn} & \textbf{mr} & \textbf{or} & \textbf{rj} & \textbf{ta} & \textbf{te} &  &  &  \\
\hline
\parbox[t]{2mm}{\multirow{13}{*}{\rotatebox[origin=c]{90}{\textbf{Actual}}}} & \textbf{as} & \textbf{1746} & 0 & 0 & 0 & 0 & 0 & 0 & 19 & 0 & 0 & 0 & 1 & 0 & 0.995 & 0.989 & 0.992 \\ \cline{2-18} 
 & \textbf{bn} & 8 & \textbf{853} & 0 & 28 & 0 & 0 & 0 & 28 & 0 & 0 & 0 & 17 & 10 & 1 & 0.904 & 0.949 \\ \cline{2-18} 
 & \textbf{bd} & 0 & 0 & \textbf{57} & 0 & 0 & 0 & 0 & 0 & 0 & 0 & 0 & 0 & 0 & 0.966 & 1 & 0.983 \\ \cline{2-18} 
 & \textbf{gu} & 0 & 0 & 0 & \textbf{567} & 0 & 0 & 0 & 0 & 0 & 0 & 0 & 0 & 1 & 0.951 & 0.998 & 0.974 \\ \cline{2-18} 
 & \textbf{hi} & 0 & 0 & 0 & 0 & \textbf{460} & 0 & 3 & 0 & 0 & 0 & 1 & 0 & 0 & 0.991 & 0.991 & 0.991 \\ \cline{2-18} 
 & \textbf{kn} & 0 & 0 & 1 & 0 & 0 & \textbf{257} & 0 & 0 & 0 & 0 & 0 & 0 & 0 & 0.996 & 0.996 & 0.996 \\ \cline{2-18} 
 & \textbf{ml} & 0 & 0 & 1 & 0 & 4 & 1 & \textbf{1119} & 1 & 0 & 0 & 4 & 0 & 0 & 0.997 & 0.99 & 0.994 \\ \cline{2-18} 
 & \textbf{mn} & 0 & 0 & 0 & 1 & 0 & 0 & 0 & \textbf{1789} & 0 & 0 & 0 & 0 & 1 & 0.973 & 0.999 & 0.986 \\ \cline{2-18} 
 & \textbf{mr} & 0 & 0 & 0 & 0 & 0 & 0 & 0 & 0 & \textbf{245} & 0 & 0 & 0 & 0 & 1 & 1 & 1 \\ \cline{2-18} 
 & \textbf{or} & 0 & 0 & 0 & 0 & 0 & 0 & 0 & 1 & 0 & \textbf{715} & 0 & 0 & 0 & 1 & 0.999 & 0.999 \\ \cline{2-18} 
 & \textbf{rj} & 0 & 0 & 0 & 0 & 0 & 0 & 0 & 0 & 0 & 0 & \textbf{912} & 0 & 0 & 0.995 & 1 & 0.997 \\ \cline{2-18} 
 & \textbf{ta} & 1 & 0 & 0 & 0 & 0 & 0 & 0 & 1 & 0 & 0 & 0 & \textbf{694} & 0 & 0.975 & 0.997 & 0.986 \\ \cline{2-18} 
 & \textbf{te} & 0 & 0 & 0 & 0 & 0 & 0 & 0 & 0 & 0 & 0 & 0 & 0 & \textbf{653} & 0.982 & 1 & 0.991 \\
\hline
\end{tabular}
\end{table}

\begin{table}[h!]
\tiny
\caption{Confusion matrix for CNN}
\centering
\label{tab: 13_category_full_cnn}
\begin{tabular}{c|c|c|c|c|c|c|c|c|c|c|c|c|c|c|c|c|c}
\hline
\multicolumn{2}{c|}{\multirow{2}{*}{}} & \multicolumn{13}{c|}{\textbf{Predicted}} & \multirow{2}{*}{\textbf{PPV}} & \multirow{2}{*}{\textbf{TPR}} & \multirow{2}{*}{\textbf{\begin{tabular}[c]{@{}c@{}}f1\\[-0.5em] Score\end{tabular}}} \\ \cline{3-15}
\multicolumn{2}{c|}{} & \textbf{as} & \textbf{bn} & \textbf{bd} & \textbf{gu} & \textbf{hi} & \textbf{kn} & \textbf{ml} & \textbf{mn} & \textbf{mr} & \textbf{or} & \textbf{rj} & \textbf{ta} & \textbf{te} &  &  &  \\
\hline
\parbox[t]{2mm}{\multirow{13}{*}{\rotatebox[origin=c]{90}{\textbf{Actual}}}} & \textbf{as} & \textbf{1744} & 0 & 1 & 0 & 0 & 3 & 1 & 13 & 0 & 0 & 0 & 4 & 0 & 0.991 & 0.988 & 0.989 \\ \cline{2-18} 
 & \textbf{bn} & 9 & \textbf{838} & 0 & 1 & 1 & 0 & 0 & 35 & 0 & 8 & 5 & 34 & 13 & 1 & 0.888 & 0.941 \\ \cline{2-18} 
 & \textbf{bd} & 0 & 0 & \textbf{57} & 0 & 0 & 0 & 0 & 0 & 0 & 0 & 0 & 0 & 0 & 0.966 & 1 & 0.983 \\ \cline{2-18} 
 & \textbf{gu} & 2 & 0 & 0 & \textbf{563} & 0 & 0 & 0 & 0 & 0 & 1 & 0 & 0 & 2 & 0.996 & 0.991 & 0.994 \\ \cline{2-18} 
 & \textbf{hi} & 0 & 0 & 0 & 0 & \textbf{452} & 0 & 10 & 0 & 0 & 0 & 2 & 0 & 0 & 0.991 & 0.974 & 0.983 \\ \cline{2-18} 
 & \textbf{kn} & 0 & 0 & 0 & 0 & 1 & \textbf{256} & 0 & 0 & 0 & 0 & 1 & 0 & 0 & 0.973 & 0.992 & 0.983 \\ \cline{2-18} 
 & \textbf{ml} & 0 & 0 & 1 & 0 & 2 & 2 & \textbf{1122} & 0 & 0 & 0 & 3 & 0 & 0 & 0.99 & 0.993 & 0.991 \\ \cline{2-18} 
 & \textbf{mn} & 1 & 0 & 0 & 0 & 0 & 0 & 0 & \textbf{1788} & 0 & 0 & 0 & 1 & 1 & 0.972 & 0.998 & 0.985 \\ \cline{2-18} 
 & \textbf{mr} & 0 & 0 & 0 & 0 & 0 & 0 & 0 & 0 & \textbf{244} & 1 & 0 & 0 & 0 & 1 & 0.996 & 0.998 \\ \cline{2-18} 
 & \textbf{or} & 0 & 0 & 0 & 0 & 0 & 0 & 0 & 1 & 0 & \textbf{715} & 0 & 0 & 0 & 0.986 & 0.999 & 0.992 \\ \cline{2-18} 
 & \textbf{rj} & 1 & 0 & 0 & 0 & 0 & 2 & 1 & 0 & 0 & 0 & \textbf{908} & 0 & 0 & 0.986 & 0.996 & 0.991 \\ \cline{2-18} 
 & \textbf{ta} & 3 & 0 & 0 & 1 & 0 & 0 & 0 & 1 & 0 & 0 & 2 & \textbf{688} & 1 & 0.946 & 0.989 & 0.967 \\ \cline{2-18} 
 & \textbf{te} & 0 & 0 & 0 & 0 & 0 & 0 & 0 & 1 & 0 & 0 & 0 & 0 & \textbf{652} & 0.975 & 0.998 & 0.986 \\
\hline
\end{tabular}
\end{table}

From Table \ref{tab: 13_category_full_Atten}, Table \ref{tab: 13_category_full_CRNN}, and Table \ref{tab: 13_category_full_cnn} it can be observed that Assamese gets confused with Manipuri; Bengali gets confused with Assamese, Manipuri, Tamil, and Telugu; and Hindi gets confused with Malayalam.

Assamese and Bengali have originated from the same language family and they share approximately the same phoneme set. However, Bengali and Tamil are from different language families but share a similar phoneme set. For example, in Bengali \textbf{cigar} is \textit{churut} and \textbf{star} is \textit{nakshatra} while \textbf{cigar} in Tamil is \textit{charuttu} and \textbf{star} in Tamil is \textit{natsattira}, which is quite similar. Similarly, Manipuri and Assamese share similar phonemes. On close study, we observed that Hindi and Malayalam have also similar phoneme sets as both languages borrowed most of the vocabulary from Sanskrit. For example, `arrogant' is \textbf{Ahankar} in Hindi and \textit{Ahankaram} in Malayalam. Similarly, \textbf{Sathyu} or commonly spoken as \textbf{Satya} in Hindi means `Truth', which is \textit{Sathyam} in Malayalam. Also, the word \textbf{Sundar} in Hindi is \textit{Sundaram} in Malayalam, which means `beautiful'.

Table \ref{tab: errors} shows the most common classification errors encountered during evaluation.

\begin{table}[h!]
\small
\caption{Most common errors}
\centering
\label{tab: errors}
\begin{tabular}{@{\extracolsep{\fill}}ccc}
\hline
Assamese & $\rightarrow$ & Manipuri  \\ 
Bengali & $\rightarrow$ & Assamese \\ 
Bengali & $\rightarrow$ & Manipuri \\ 
Bengali & $\rightarrow$ & Tamil \\ 
Hindi & $\rightarrow$ & Malayalam \\
\hline
\end{tabular}
\end{table}

\subsection{Result on same language families on Indian Language Dataset}
\label{subsec:Results_same_lang_family}
A deeper study into these 13 Indian languages led us to define five clusters of languages based on their phonetic similarity. Cluster internal languages are phonetically similar, close, and geographically contiguous, hence difficult to differentiate.
\begin{itemize}
    \vspace{-1em}
    \item \textit{\textbf{Cluster 1}: Assamese, Bengali, Odia}
    \item \textit{\textbf{Cluster 2}: Gujarati, Hindi, Marathi, Rajasthani}
    \item \textit{\textbf{Cluster 3}: Kannada, Malayalam, Tamil, Telugu}
    \item \textit{\textbf{Cluster 4}: Bodo}
    \item \textit{\textbf{Cluster 5}: Manipuri}
\end{itemize}

Bodo and Manipuri are phonetically very much distant from any of the rest of the languages, thus they form singleton clusters. We carried out separate experiments for the identification of the cluster internal languages for Cluster 1, 2 and 3, and the experimental results are presented in Table \ref{tab: res_lang_group}.

\begin{table}[h!]
\tiny
\caption{Experimental Results of LID for close languages.}
\centering
\label{tab: res_lang_group}
\begin{tabular}{c|c|c|c|c|c|c|c|c|c|c|c|c|c}
\hline
\multirow{2}{*}{\rotatebox[origin=c]{90}{\textbf{Cluster}}} & \multirow{2}{*}{\rotatebox[origin=c]{90}{\textbf{Language}}} & \multicolumn{4}{c|}{\textbf{\begin{tabular}[c]{@{}c@{}}CRNN \\[-0.5em] with \\[-0.5em] Attention\end{tabular} }} & \multicolumn{4}{c|}{\textbf{CRNN}} & \multicolumn{4}{c}{\textbf{CNN}} \\ \cline{3-14} 
    & & \textbf{PPV} & \textbf{TPR} & \textbf{\begin{tabular}[c]{@{}c@{}}f1\\[-0.5em] Score\end{tabular}} & \textbf{Accuracy} & \textbf{PPV} & \textbf{TPR} & \textbf{\begin{tabular}[c]{@{}c@{}}f1\\[-0.5em] Score\end{tabular}} & \textbf{Accuracy} & \textbf{PPV} & \textbf{TPR} & \textbf{\begin{tabular}[c]{@{}c@{}}f1\\[-0.5em] Score\end{tabular}} & \textbf{Accuracy}\\
    \hline
    \multirow{3}{*}{\textbf{1}} & \textbf{as} & 0.962 & 1 & 0.981 & \multirow{3}{*}{\textbf{0.98}} & 0.953 & 1 & 0.976 & \multirow{3}{*}{\textbf{0.974}} & 0.953 & 1 & 0.976 & \multirow{3}{*}{\textbf{0.971}} \\ \cline{2-5} \cline{7-9} \cline{11-13}
    & \textbf{bn} & 1 & 0.926 & 0.961 &  & 1 & 0.907 & 0.951 &  & 1 & 0.894 & 0.944 &  \\ \cline{2-5} \cline{7-9} \cline{11-13}
    & \textbf{or} & 1 & 1 & 1 &  & 1 & 1 & 1 &  & 0.982 & 1 & 0.991 &  \\ \hline\hline
    \multirow{4}{*}{\textbf{2}} & \textbf{gu} & 1 & 0.998 & 0.999 & \multirow{4}{*}{\textbf{0.999}} & 1 & 0.998 & 0.999 & \multirow{4}{*}{\textbf{0.999}} & 1 & 0.993 & 0.996 & \multirow{4}{*}{\textbf{0.996}} \\ \cline{2-5} \cline{7-9} \cline{11-13}
    & \textbf{hi} & 1 & 1 & 1 &  & 1 & 0.998 & 0.999 &  & 0.991 & 0.996 & 0.994 &  \\ \cline{2-5} \cline{7-9} \cline{11-13}
    & \textbf{mr} & 1 & 1 & 1 &  & 1 & 1 & 1 &  & 1 & 0.996 & 0.998 &  \\ \cline{2-5} \cline{7-9} \cline{11-13}
    & \textbf{rj} & 0.999 & 1 & 0.999 &  & 0.998 & 1 & 0.999 &  & 0.995 & 0.998 & 0.996 &  \\ \hline\hline
    \multirow{4}{*}{\textbf{3}} & \textbf{kn} & 1 & 0.996 & 0.998 & \multirow{4}{*}{\textbf{0.999}} & 1 & 1 & 1 & \multirow{4}{*}{\textbf{1}} & 0.992 & 0.988 & 0.99 & \multirow{4}{*}{\textbf{0.996}} \\ \cline{2-5} \cline{7-9} \cline{11-13}
    & \textbf{ml} & 0.999 & 1 & 0.999 &  & 1 & 1 & 1 &  & 0.996 & 0.996 & 0.996 &  \\ \cline{2-5} \cline{7-9} \cline{11-13}
    & \textbf{ta} & 1 & 1 & 1 &  & 1 & 1 & 1 &  & 0.996 & 0.997 & 0.996 &  \\ \cline{2-5} \cline{7-9} \cline{11-13}
    & \textbf{te} & 1 & 1 & 1 &  & 1 & 1 & 1 &  & 0.995 & 0.997 & 0.996 &  \\
\hline
\end{tabular}
\end{table}

It can be clearly observed from Table \ref{tab: res_lang_group} that both CRNN framework and CRNN with Attention provide competitive results for every language cluster. For \textbf{cluster-1} CRNN framework and CRNN with Attention provides an accuracy of 0.98/0.974, for \textbf{cluster-2} 0.999/0.999, and for \textbf{cluster-3} 0.999/1, respectively. CNN framework also provides comparable results to the other two frameworks.

\begin{table}[h!]
\tiny
\caption{Confusion matrix for Cluster 1}
\vspace{-1.5em}
\centering
\label{tab: lang_family_AsBnOr}
\begin{tabular}{c|c|c|c|c|c|c|c}
\hline
\multicolumn{8}{c}{\textbf{CRNN and Attention}} \\ \hline
\multicolumn{2}{c|}{\multirow{2}{*}{\textbf{}}} &
\multicolumn{3}{c|}{\textbf{Predicted}} &
\multirow{2}{*}{\textbf{PPV}} &
\multirow{2}{*}{\textbf{TPR}} &
\multirow{2}{*}{\textbf{\begin{tabular}[c]{@{}c@{}}f1\\[-0.5em] Score\end{tabular}}} \\ \cline{3-5}
\multicolumn{2}{c|}{} &
\textbf{as} &
\textbf{bn} &
\textbf{or} &
    &
    & \\
\hline
\multirow{3}{*}{\textbf{Actual}} &
\textbf{as} &
\textbf{1766} &
    0 &
    0 &
    0.962 &
    1 &
    0.981 \\ \cline{2-8} 
    &
    \textbf{bn} &
    70 &
    \textbf{874} &
    0 &
    1 &
    0.926 &
    0.961 \\ \cline{2-8} 
    &
    \textbf{or} &
    0 &
    0 &
    \textbf{716} &
    1 &
    1 &
    1  \\
\hline\hline
\multicolumn{8}{c}{\textbf{CRNN}} \\ \hline
\multicolumn{2}{c|}{\multirow{2}{*}{\textbf{}}} &
\multicolumn{3}{c|}{\textbf{Predicted}} &
\multirow{2}{*}{\textbf{PPV}} &
\multirow{2}{*}{\textbf{TPR}} &
\multirow{2}{*}{\textbf{\begin{tabular}[c]{@{}c@{}}f1\\[-0.5em] Score\end{tabular}}} \\ \cline{3-5}
\multicolumn{2}{c|}{} &
\textbf{as} &
\textbf{bn} &
\textbf{or} &
    &
    & \\
    \hline
    \multirow{3}{*}{\textbf{Actual}} &
    \textbf{as} &
    \textbf{1766} &
    0 &
    0 &
    0.953 &
    1 &
    0.976 \\ \cline{2-8} 
    &
    \textbf{bn} &
    88 &
    \textbf{856} &
    0 &
    1 &
    0.907 &
    0.951 \\ \cline{2-8} 
    &
    \textbf{or} &
    0 &
    0 &
    \textbf{716} &
    1 &
    1 &
    1 \\
    \hline\hline
    \multicolumn{8}{c}{\textbf{CNN}} \\ \hline
    \multicolumn{2}{c|}{\multirow{2}{*}{\textbf{}}} &
    \multicolumn{3}{c|}{\textbf{Predicted}} &
    \multirow{2}{*}{\textbf{PPV}} &
    \multirow{2}{*}{\textbf{TPR}} &
    \multirow{2}{*}{\textbf{\begin{tabular}[c]{@{}c@{}}f1\\[-0.5em] Score\end{tabular}}} \\ \cline{3-5}
    \multicolumn{2}{c|}{} &
    \textbf{as} &
    \textbf{bn} &
    \textbf{or} &
    &
    &
    \\ \hline
    \multirow{3}{*}{\textbf{Actual}} &
    \textbf{as} &
    \textbf{1766} &
    0 &
    0 &
    0.953 &
    1 &
    0.976 \\ \cline{2-8} 
    &
    \textbf{bn} &
    87 &
    \textbf{844} &
    13 &
    1 &
    0.894 &
    0.944 \\ \cline{2-8} 
    &
    \textbf{or} &
    0 &
    0 &
    \textbf{716} &
    0.982 &
    1 &
    0.991 \\ \hline
    \end{tabular}
\end{table}

\begin{table}[h!]
\tiny
 \begin{minipage}{.5\linewidth}
\caption{Confusion matrix for Cluster 2}
\vspace{-1.5em}
\centering
\label{tab: lang_family_GujHinMarRaj}
\begin{tabular}{c|c|c|c|c|c|c|c|c}
    \hline
    \multicolumn{9}{c}{\textbf{CRNN and Attention}} \\ \hline
\multicolumn{2}{c|}{\multirow{2}{*}{\textbf{}}} &
  \multicolumn{4}{c|}{\textbf{Predicted}} &
  \multirow{2}{*}{\textbf{PPV}} &
  \multirow{2}{*}{\textbf{TPR}} &
  \multirow{2}{*}{\textbf{\begin{tabular}[c]{@{}c@{}}f1\\[-0.5em] Score\end{tabular}}} \\ \cline{3-6}
\multicolumn{2}{c|}{} &
  \textbf{gu} &
  \textbf{hi} &
  \textbf{mr} &
  \textbf{rj} &
   &
   &
   \\ \hline
    \multirow{4}{*}{\textbf{Actual}} &
  \textbf{gu} &
  \textbf{567} &
  0 &
  0 &
  1 &
  1 &
  0.998 &
  0.999 \\ \cline{2-9} 
 &
  \textbf{hi} &
  0 &
  \textbf{464} &
  0 &
  0 &
  1 &
  1 &
  1 \\ \cline{2-9} 
 &
  \textbf{mr} &
  0 &
  0 &
  \textbf{245} &
  0 &
  1 &
  1 &
  1 \\ \cline{2-9} 
 &
  \textbf{rj} &
  0 &
  0 &
  0 &
  \textbf{912} &
  0.999 &
  1 &
  0.999 \\ \hline\hline
\multicolumn{9}{c}{\textbf{CRNN}} \\ \hline
\multicolumn{2}{c|}{\multirow{2}{*}{\textbf{}}} &
  \multicolumn{4}{c|}{\textbf{Predicted}} &
  \multirow{2}{*}{\textbf{PPV}} &
  \multirow{2}{*}{\textbf{TPR}} &
  \multirow{2}{*}{\textbf{\begin{tabular}[c]{@{}c@{}}f1\\[-0.5em] Score\end{tabular}}} \\ \cline{3-6}
\multicolumn{2}{c|}{} &
  \textbf{gu} &
  \textbf{hi} &
  \textbf{mr} &
  \textbf{rj} &
   &
   &
   \\ \hline
\multirow{4}{*}{\textbf{Actual}} &
  \textbf{gu} &
  \textbf{567} &
  0 &
  0 &
  1 &
  1 &
  0.998 &
  0.999 \\ \cline{2-9} 
 &
  \textbf{hi} &
  0 &
  \textbf{463} &
  0 &
  1 &
  1 &
  0.998 &
  0.999 \\ \cline{2-9} 
 &
  \textbf{mr} &
  0 &
  0 &
  \textbf{245} &
  0 &
  1 &
  1 &
  1 \\ \cline{2-9} 
 &
  \textbf{rj} &
  0 &
  0 &
  0 &
  \textbf{912} &
  0.998 &
  1 &
  0.999 \\ \hline\hline
\multicolumn{9}{c}{\textbf{CNN}} \\ \hline
\multicolumn{2}{c|}{\multirow{2}{*}{\textbf{}}} &
  \multicolumn{4}{c|}{\textbf{Predicted}} &
  \multirow{2}{*}{\textbf{PPV}} &
  \multirow{2}{*}{\textbf{TPR}} &
  \multirow{2}{*}{\textbf{\begin{tabular}[c]{@{}c@{}}f1\\[-0.5em] Score\end{tabular}}} \\ \cline{3-6}
\multicolumn{2}{c|}{} &
  \textbf{gu} &
  \textbf{hi} &
  \textbf{mr} &
  \textbf{rj} &
   &
   &
   \\ \hline
\multirow{4}{*}{\textbf{Actual}} &
  \textbf{gu} &
  \textbf{564} &
  2 &
  0 &
  2 &
  1 &
  0.993 &
  0.996 \\ \cline{2-9} 
 &
  \textbf{hi} &
  0 &
  \textbf{462} &
  0 &
  2 &
  0.991 &
  0.996 &
  0.994 \\ \cline{2-9} 
 &
  \textbf{mr} &
  0 &
  0 &
  \textbf{244} &
  1 &
  1 &
  0.996 &
  0.998 \\ \cline{2-9} 
 &
  \textbf{rj} &
  0 &
  2 &
  0 &
  \textbf{910} &
  0.995 &
  0.998 &
  0.996 \\ \hline
    \end{tabular}
\end{minipage}%
\begin{minipage}{.5\linewidth}
\caption{Confusion matrix for Cluster 3}
\vspace{-1.5em}
\centering
\label{tab: lang_family_KanMalTamTel}
\begin{tabular}{c|c|c|c|c|c|c|c|c}
    \hline
    \multicolumn{9}{c}{\textbf{CRNN and Attention}} \\ \hline
\multicolumn{2}{c|}{\multirow{2}{*}{\textbf{}}} &
  \multicolumn{4}{c|}{\textbf{Predicted}} &
  \multirow{2}{*}{\textbf{PPV}} &
  \multirow{2}{*}{\textbf{TPR}} &
  \multirow{2}{*}{\textbf{\begin{tabular}[c]{@{}c@{}}f1\\[-0.5em] Score\end{tabular}}} \\ \cline{3-6}
\multicolumn{2}{c|}{} &
  \textbf{kn} &
  \textbf{ml} &
  \textbf{ta} &
  \textbf{te} &
   &
   &
   \\ \hline
\multirow{4}{*}{\textbf{Actual}} &
  \textbf{kn} &
  \textbf{257} &
  1 &
  0 &
  0 &
  1 &
  0.996 &
  0.998 \\ \cline{2-9} 
 &
  \textbf{ml} &
  0 &
  \textbf{1130} &
  0 &
  0 &
  0.999 &
  1 &
  0.999 \\ \cline{2-9} 
 &
  \textbf{ta} &
  0 &
  0 &
  \textbf{696} &
  0 &
  1 &
  1 &
  1 \\ \cline{2-9} 
 &
  \textbf{te} &
  0 &
  0 &
  0 &
  \textbf{653} &
  1 &
  1 &
  1  \\ \hline\hline
\multicolumn{9}{c}{\textbf{CRNN}} \\ \hline
\multicolumn{2}{c|}{\multirow{2}{*}{\textbf{}}} &
  \multicolumn{4}{c|}{\textbf{Predicted}} &
  \multirow{2}{*}{\textbf{PPV}} &
  \multirow{2}{*}{\textbf{TPR}} &
  \multirow{2}{*}{\textbf{\begin{tabular}[c]{@{}c@{}}f1\\[-0.5em] Score\end{tabular}}} \\ \cline{3-6}
\multicolumn{2}{c|}{} &
  \textbf{kn} &
  \textbf{ml} &
  \textbf{ta} &
  \textbf{te} &
   &
   &
   \\ \hline
\multirow{4}{*}{\textbf{Actual}} &
  \textbf{kn} &
  \textbf{258} &
  0 &
  0 &
  0 &
  1 &
  1 &
  1 \\ \cline{2-9} 
 &
  \textbf{ml} &
  0 &
  \textbf{1130} &
  0 &
  0 &
  1 &
  1 &
  1 \\ \cline{2-9} 
 &
  \textbf{ta} &
  0 &
  0 &
  \textbf{696} &
  0 &
  1 &
  1 &
  1 \\ \cline{2-9} 
 &
  \textbf{te} &
  0 &
  0 &
  0 &
  \textbf{653} &
  1 &
  1 &
  1  \\ \hline\hline
\multicolumn{9}{c}{\textbf{CNN}} \\ \hline
\multicolumn{2}{c|}{\multirow{2}{*}{\textbf{}}} &
  \multicolumn{4}{c|}{\textbf{Predicted}} &
  \multirow{2}{*}{\textbf{PPV}} &
  \multirow{2}{*}{\textbf{TPR}} &
  \multirow{2}{*}{\textbf{\begin{tabular}[c]{@{}c@{}}f1\\[-0.5em] Score\end{tabular}}} \\ \cline{3-6}
\multicolumn{2}{c|}{} &
  \textbf{kn} &
  \textbf{ml} &
  \textbf{ta} &
  \textbf{te} &
   &
   &
   \\ \hline
\multirow{4}{*}{\textbf{Actual}} &
  \textbf{kn} &
  \textbf{255} &
  3 &
  0 &
  0 &
  0.992 &
  0.988 &
  0.99 \\ \cline{2-9} 
 &
  \textbf{ml} &
  1 &
  \textbf{1125} &
  2 &
  2 &
  0.996 &
  0.996 &
  0.996 \\ \cline{2-9} 
 &
  \textbf{ta} &
  1 &
  0 &
  \textbf{694} &
  1 &
  0.996 &
  0.997 &
  0.996 \\ \cline{2-9} 
 &
  \textbf{te} &
  0 &
  1 &
  1 &
  \textbf{651} &
  0.995 &
  0.997 &
  0.996 \\ \hline
    \end{tabular}
    \end{minipage}
\end{table}

Table \ref{tab: lang_family_AsBnOr}, Table \ref{tab: lang_family_GujHinMarRaj} and Table \ref{tab: lang_family_KanMalTamTel} presents the confusion matrix for cluster 1, cluster 2, and cluster 3, respectively.
From Table \ref{tab: lang_family_AsBnOr}, we observed that Bengali gets confused with Assamese and Odia, which is quite expected since these two languages are spoken in neighbouring states and both of them share almost the same phonemes. For example, in Odia \textbf{rice} is pronounced as \textit{bhata} whereas in Bengali pronounced as \textit{bhat}, similarly \textbf{fish} in odia as \textit{machha} whereas in Bengali it is \textit{machh}. Both CRNN and CRNN with Attention perform well to discriminate between Bengali and Odia.
It can be observed from Table \ref{tab: lang_family_KanMalTamTel} that CNN creates a lot of confusion when discriminating between these four languages. Both CRNN and CRNN with Attention prove to be better at discriminating among these languages. From the results in Table \ref{tab: res_lang_group}, \ref{tab: lang_family_AsBnOr}, \ref{tab: lang_family_GujHinMarRaj} and \ref{tab: lang_family_KanMalTamTel}, it is pretty clear that CRNN (Bi-Directional LSTM over CNN) and CRNN with Attention are more effective for Indian language identification and they perform almost at par. Another important observation is that it is harder to classify the languages in cluster 1 than in the other two clusters.

\subsection{Results on European Language}
\label{subsec:results_eu_data}
We evaluated our model in two environments -- No Noise and White Noise. According to our intuition, in real-life scenarios during prediction of language chances of capturing background noise of chatter and other sounds may happen. For the white noise evaluation setup, we mixed white noise into each test sample which has an audible solid presence but retains the identity of the language.

\begin{table}[h!]
\small
\caption{Comparative evaluation results (in terms of Accuracy) of our model and the model of \citet{Bartz2017} on the EU dataset}
\centering
\label{tab: eval}
\begin{tabular}{ccc}
    \hline
        & \textbf{No Noise} & \textbf{White Noise} \\
    \hline
    CRNN \citep{Bartz2017}  & 0.91  & 0.63\\\hdashline
    Inception-v3 CRNN \citep{Bartz2017}   & 0.96   & 0.91\\\hdashline
    CNN    & 0.948 & 0.871\\\hdashline
    CRNN   & \textbf{0.967}    & \textbf{0.912}\\\hdashline
    CRNN with Attention    & 0.966 & 0.888 \\
    \hline
    \end{tabular}
\end{table}

Table \ref{tab: eval} compares the results of our models on the EU dataset with state-of-the-art models presented by \citet{Bartz2017}. The model proposed by \citet{Bartz2017} consists of CRNN and uses Google's Inception-v3 framework \citep{Szegedy2016}. The feature extractor performs convolutional operations on the input image through multiple stages, resulting in the production of a feature map that possesses a height of one. The feature map is partitioned horizontally along the x-axis, and each partition is employed as a temporal unit for the subsequent Bidirectional LSTM network. The network employs a total of five convolutional layers, with each layer being succeeded by the ReLU activation function, Batch Normalization, and $2\times2$ max pooling with a stride of 2. The convolutional layers in question are characterized by their respective kernel sizes and the number of filters, which are as follows: ($7\times7$, 16), ($5\times5$, 32), ($3\times3$, 64), ($3\times3$, 128), and ($3\times3$, 256). The Bidirectional LSTM model comprises a pair of individual LSTM models, each with 256 output units. The concatenation of the two outputs is transformed into a 512-dimensional vector, which is then input into a fully-connected layer. The layer has either 4 or 6 output units, which function as the classifier. They experimented in four different environments -- No Noise, White Noise, Cracking Noise, and Background Noise. All our evaluation results are rounded to 3 digits after the decimal point.

The CNN model failed to achieve competitive results; it provided an accuracy of 0.948/0.871 in No Noise/White Noise. In the CRNN framework, our model provides an accuracy of 0.967/0.912 on the No Noise/White Noise scenario outperforming the state-of-the-art results of \citet{Bartz2017}. Use of Attention improves over the Inception-v3 CRNN in the No Noise scenario, however, it does not perform well on White Noise.

\subsection{Ablation Studies}
\label{subsec:ablation}
\subsubsection{Convolution Kernel Size}
\label{subsubsec:cnn}

To study the effect of kernel sizes in the convolution layers, we sweep the kernel size with 3, 7, 17, 32, and 65  of the models. We found that performance decreases with larger kernel sizes, as shown in Table \ref{tab: kernel_size}. On comparing the accuracy up to the second decimal place kernel size 3 performs better than the rest.

\begin{table}[h!]
\small
\caption{Ablation study on convolution kernel sizes}
\centering
\label{tab: kernel_size}
\begin{tabular}{cc}
\hline
\textbf{Kernel size} & \textbf{Accuracy} \\ \hline
3                    & 98.7\%            \\ 
7                    & 98.68\%           \\ 
17                   & 98.65\%           \\ 
32                   & 98.13\%           \\ 
65                   & 93.56\%           \\ \hline
\end{tabular}
\end{table}

\subsubsection{Automatic Class Weight vs Manual Class Weight}
Balancing the data using class weights gives better accuracy for CRNN with Attention (98.7\%) and CRNN (98.7\%), compared to CNN (98.3\%) shown in Table \ref{tab: res_lang_class_weight}. We study the efficacy of the frameworks by manually balancing the datasets using 100 samples, 200 samples, and 571 samples drawn randomly from the dataset and the results of these experiments are presented in Table \ref{tab: res_lang_class_100}, Table \ref{tab: res_lang_class_200} and Table \ref{tab: res_lang_class_571}, respectively.

\begin{table}[h!]
\tiny 
\caption{Experimental Results for Manually Balancing the Samples for each category to 100.}
\centering
\label{tab: res_lang_class_100}
\begin{tabular}{ c|c|c|c|c|c|c|c|c|c|c|c|c }
\hline
\multirow{2}{*}{\textbf{Language}} & \multicolumn{4}{c|}{\textbf{CRNN with Attention}} & \multicolumn{4}{c|}{\textbf{CRNN}} & \multicolumn{4}{c}{\textbf{CNN}} \\ \cline{2-13} 
& \textbf{PPV} & \textbf{TPR} & \textbf{\begin{tabular}[c]{@{}c@{}}f1\\ Score\end{tabular}} & \textbf{Accuracy} & \textbf{PPV} & \textbf{TPR} & \textbf{\begin{tabular}[c]{@{}c@{}}f1\\ Score\end{tabular}} & \textbf{Accuracy} & \textbf{PPV} & \textbf{TPR} & \textbf{\begin{tabular}[c]{@{}c@{}}f1\\[-0.5em] Score\end{tabular}} & \textbf{Accuracy} \\
\hline
\textbf{as} & 0.766 & 0.72 & 0.742 & \multirow{13}{*}{\textbf{0.883}} & 0.839 & 0.94 & 0.887 & \multirow{13}{*}{\textbf{0.932}} & 0.617 & 0.58 & 0.598 & \multirow{13}{*}{\textbf{0.72}} \\ \cline{1-4} \cline{6-8} \cline{10-12}
\textbf{bn} & 0.875 & 0.7 & 0.778 &  & 0.957 & 0.9 & 0.928 &  & 0.816 & 0.8 & 0.808 &  \\ \cline{1-4} \cline{6-8} \cline{10-12}
\textbf{bd} & 1 & 1 & 1 &  & 0.962 & 1 & 0.98 &  & 0.843 & 0.86 & 0.851 &  \\ \cline{1-4} \cline{6-8} \cline{10-12}
\textbf{gu} & 0.943 & 1 & 0.971 &  & 1 & 0.98 & 0.99 &  & 0.731 & 0.76 & 0.745 &  \\ \cline{1-4} \cline{6-8} \cline{10-12}
\textbf{hi} & 0.959 & 0.94 & 0.95 &  & 0.957 & 0.9 & 0.928 &  & 0.778 & 0.7 & 0.737 &  \\ \cline{1-4} \cline{6-8} \cline{10-12}
\textbf{kn} & 0.961 & 0.98 & 0.97 &  & 0.94 & 0.94 & 0.94 &  & 0.725 & 0.74 & 0.733 &  \\ \cline{1-4} \cline{6-8} \cline{10-12}
\textbf{ml} & 0.958 & 0.92 & 0.939 &  & 0.923 & 0.96 & 0.941 &  & 0.774 & 0.82  & 0.796 &  \\ \cline{1-4} \cline{6-8} \cline{10-12}
\textbf{mn} & 0.878 & 0.72 & 0.791 &  & 0.935 & 0.86 & 0.896 &  & 0.691 & 0.76 & 0.724 &  \\ \cline{1-4} \cline{6-8} \cline{10-12}
\textbf{mr} & 0.906 & 0.96 & 0.932 &  & 0.98 & 0.96 & 0.97 &  & 0.857 & 0.84 & 0.848 &   \\ \cline{1-4} \cline{6-8} \cline{10-12}
\textbf{or} & 0.959 & 0.94 & 0.949 &  & 0.943 & 1 & 0.971 &  & 0.811 & 0.86 & 0.835 &   \\ \cline{1-4} \cline{6-8} \cline{10-12}
\textbf{rj} & 0.782 & 0.86 & 0.819 &  & 0.894 & 0.84 & 0.866 &  & 0.605 & 0.52 & 0.559 &  \\ \cline{1-4} \cline{6-8} \cline{10-12}
\textbf{ta} & 0.677 & 0.88 & 0.765 &  & 0.898 & 0.88 & 0.889 &  & 0.564 & 0.62 & 0.590 &   \\ \cline{1-4} \cline{6-8} \cline{10-12}
\textbf{te} & 0.878 & 0.86 & 0.869 &  & 0.906 & 0.96 & 0.932 &  & 0.532 & 0.5 & 0.515 &  \\
\hline
\end{tabular}
\end{table}

\begin{table}[h!]
\tiny 
\caption{Experimental Results for Manually Balancing the Samples for each Category to 200.}
\centering
\label{tab: res_lang_class_200}
\begin{tabular}{c|c|c|c|c|c|c|c|c|c|c|c|c }
\hline
\multirow{2}{*}{\textbf{Language}} & \multicolumn{4}{c|}{\textbf{CRNN with Attention}} & \multicolumn{4}{c|}{\textbf{CRNN}} & \multicolumn{4}{c}{\textbf{CNN}} \\ \cline{2-13} 
    & \textbf{PPV} & \textbf{TPR} & \textbf{\begin{tabular}[c]{@{}c@{}}f1\\[-0.5em] Score\end{tabular}} & \textbf{Accuracy} & \textbf{PPV} & \textbf{TPR} & \textbf{\begin{tabular}[c]{@{}c@{}}f1\\[-0.5em] Score\end{tabular}} & \textbf{Accuracy} & \textbf{PPV} & \textbf{TPR} & \textbf{\begin{tabular}[c]{@{}c@{}}f1\\[-0.5em] Score\end{tabular}} & \textbf{Accuracy}  \\
    \hline
    \textbf{as} & 0.941 & 0.96 & 0.95 & \multirow{13}{*}{\textbf{0.975}} & 1 & 0.94 & 0.969 & \multirow{13}{*}{\textbf{0.971}} & 0.8 & 0.88 & 0.838 & \multirow{13}{*}{\textbf{0.883}} \\ \cline{1-4} \cline{6-8} \cline{10-12}
    \textbf{bn} & 0.909 & 1 & 0.952 &  & 1 & 0.96 & 0.98 &  & 0.92 & 0.92 & 0.92 &  \\ \cline{1-4} \cline{6-8} \cline{10-12}
    \textbf{bd} & 0.98 & 0.96 & 0.97 &  & 0.98 & 0.98 & 0.98 &  & 0.94 & 0.94 & 0.94 &  \\ \cline{1-4} \cline{6-8} \cline{10-12}
    \textbf{gu} & 1 & 1 & 1 &  & 1 & 1 & 1 &  & 0.918 & 0.9 & 0.909 &  \\ \cline{1-4} \cline{6-8} \cline{10-12}
    \textbf{hi} & 1 & 0.98 & 0.99 &  & 1 & 0.98 & 0.99 &  & 0.956 & 0.86 & 0.905 &  \\ \cline{1-4} \cline{6-8} \cline{10-12}
    \textbf{kn} & 1 & 0.98 & 0.99 &  & 1 & 0.98 & 0.99 &  & 0.878 & 0.86 & 0.869 &  \\ \cline{1-4} \cline{6-8} \cline{10-12}
    \textbf{ml}  & 0.962 & 1 & 0.98 &  & 0.893 & 1 & 0.943 &  & 0.896 & 0.86 & 0.878 &  \\ \cline{1-4} \cline{6-8} \cline{10-12}
    \textbf{mn} & 0.979 & 0.92 & 0.948 &  & 0.907 & 0.98 & 0.942 &  & 0.754 & 0.92 & 0.829 &   \\ \cline{1-4} \cline{6-8} \cline{10-12}
    \textbf{mr} & 0.98 & 0.98 & 0.98 &  & 0.98 & 0.96 & 0.97 &  & 0.956 & 0.86 & 0.905 &   \\ \cline{1-4} \cline{6-8} \cline{10-12}
    \textbf{or}  & 0.98 & 1 & 0.99 &  & 1 & 1 & 1 &  & 0.941 & 0.96 & 0.95 &  \\ \cline{1-4} \cline{6-8} \cline{10-12}
    \textbf{rj} & 0.96 & 0.96 & 0.96 &  & 1 & 0.96 & 0.98 &  & 0.86 & 0.86 & 0.86 &  \\ \cline{1-4} \cline{6-8} \cline{10-12}
    \textbf{ta} & 1 & 0.96 & 0.98 &  & 0.904 & 0.94 & 0.922 &  & 0.784 & 0.8 & 0.792 &  \\ \cline{1-4} \cline{6-8} \cline{10-12}
    \textbf{te} & 1 & 0.98 & 0.99 &  & 0.979 & 0.94 & 0.959 &  & 0.935 & 0.86 & 0.896 &  \\
\hline
\end{tabular}
\end{table}

\begin{table}[h!]
\tiny 
\caption{Experimental Results for Manually Balancing the Samples for each category to 571.}
\centering
\label{tab: res_lang_class_571}
\begin{tabular}{c|c|c|c|c|c|c|c|c|c|c|c|c }
    \hline
    \multirow{2}{*}{\textbf{Language}} & \multicolumn{4}{c|}{\textbf{CRNN with Attention}} & \multicolumn{4}{c|}{\textbf{CRNN}} & \multicolumn{4}{c}{\textbf{CNN}} \\ \cline{2-13} 
    & \textbf{PPV} & \textbf{TPR} & \textbf{\begin{tabular}[c]{@{}c@{}}f1\\[-0.5em] Score\end{tabular}} & \textbf{Accuracy} & \textbf{PPV} & \textbf{TPR} & \textbf{\begin{tabular}[c]{@{}c@{}}f1\\[-0.5em] Score\end{tabular}} & \textbf{Accuracy} & \textbf{PPV} & \textbf{TPR} & \textbf{\begin{tabular}[c]{@{}c@{}}f1\\[-0.5em] Score\end{tabular}} & \textbf{Accuracy}  \\
    \hline
    \textbf{as} & 1 & 1 & 1 & \multirow{13}{*}{\textbf{0.988}} & 0.983 & 0.983 & 0.983 & \multirow{13}{*}{\textbf{0.985}} & 0.967 & 1 & 0.983 & \multirow{13}{*}{\textbf{0.945}} \\ \cline{1-4} \cline{6-8} \cline{10-12}
    \textbf{bn} & 1 & 1 & 1 &  & 1 & 1 & 1 &  & 0.983 & 1 & 0.991 &  \\ \cline{1-4} \cline{6-8} \cline{10-12}
    \textbf{bd} & 1 & 1 & 1 &  & 1 & 1 & 1 &  & 1 & 1 & 1 &  \\ \cline{1-4} \cline{6-8} \cline{10-12}
    \textbf{gu} & 1 & 1 & 1 &  & 0.983 & 1 & 0.991 &  & 0.982 & 0.931 & 0.956 &  \\ \cline{1-4} \cline{6-8} \cline{10-12}
    \textbf{hi} & 0.983 & 0.983 & 0.983 &  & 1 & 1 & 1 &  & 0.893 & 0.862 & 0.877 &  \\ \cline{1-4} \cline{6-8} \cline{10-12}
    \textbf{kn} & 1 & 1 & 1 &  & 1 & 1 & 1 &  & 0.903 & 0.966 & 0.933 &  \\ \cline{1-4} \cline{6-8} \cline{10-12}
    \textbf{ml} & 1 & 0.966 & 0.982 &  & 0.983 & 1 & 0.991 &  & 0.914 & 0.914 & 0.914 &  \\ \cline{1-4} \cline{6-8} \cline{10-12}
    \textbf{mn} & 0.983 & 1 & 0.991 &  & 1 & 0.983 & 0.991 &  & 0.931 & 0.931 & 0.931 &  \\ \cline{1-4} \cline{6-8} \cline{10-12}
    \textbf{mr} & 1 & 1 & 1 &  & 0.982 & 1 & 0.991 &  & 0.965 & 0.982 & 0.973 &  \\ \cline{1-4} \cline{6-8} \cline{10-12}
    \textbf{or} & 1 & 1 & 1 &  & 1 & 0.983 & 0.991 &  & 1 & 0.966 & 0.982 &  \\ \cline{1-4} \cline{6-8} \cline{10-12}
    \textbf{rj} & 0.919 & 0.983 & 0.95 &  & 0.918 & 0.966 & 0.941 &  & 0.9 & 0.931 & 0.915 &  \\ \cline{1-4} \cline{6-8} \cline{10-12}
    \textbf{ta} & 0.964 & 0.931 & 0.947 &  & 0.964 & 0.914 & 0.938 &  & 0.879 & 0.879 & 0.879 &  \\ \cline{1-4} \cline{6-8} \cline{10-12}
    \textbf{te} & 1 & 0.983 & 0.991 &  & 1 & 0.983 & 0.991 &  & 0.982 & 0.931 & 0.956 &  \\
    \hline
    \end{tabular}
  \end{table}

The objective of the study was to observe the performance of the frameworks in increasing the sample size. Since the Bodo language has the minimum data (571 samples) among all the languages in the dataset, we performed our experiments on 571 samples.

A comparison of the results in Table \ref{tab: res_lang_class_100}, Table \ref{tab: res_lang_class_200}, and Table \ref{tab: res_lang_class_571} reveals the following observations.
\begin{itemize}
    \vspace{-1em}
    \item All the models perform consistently better with more training data.
    \item CRNN and CRNN with attention perform consistently better than CNN.
    \item CRNN is less data hungry among the 3 models and it performs the best in the lowest data scenario.
    \vspace{-0.5em}
\end{itemize}
\begin{figure}[h!]
\centering
\includegraphics[scale=.75]{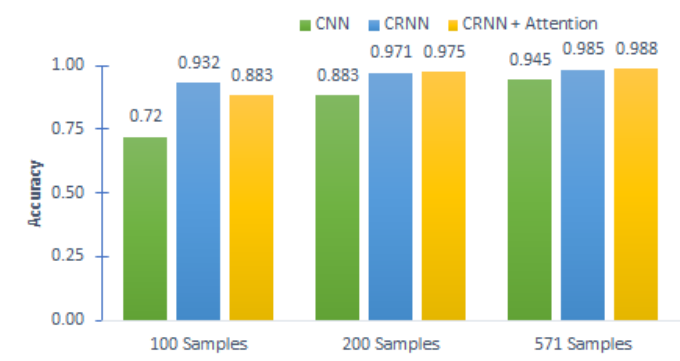}
\caption{Comparison of model results for varying dataset size.}
\label{fig:res_model}
\end{figure}

Figure \ref{fig:res_model} graphically shows the performance improvement over increasing data samples. The confusion matrices for the three frameworks for the 3 datasets are presented in Table  \ref{tab: 13_category_100_cnn}, \ref{tab: 13_category_200_cnn}, \ref{tab: 13_category_571_cnn}, \ref{tab: 13_category_100_CRNN}, \ref{tab: 13_category_200_CRNN}, \ref{tab: 13_category_571_CRNN}, \ref{tab: 13_category_100_att},
\ref{tab: 13_category_200_att},
\ref{tab: 13_category_571_att} in the Appendix.

\subsubsection{Additional performance and parameter size analysis of our frameworks}

\begin{table}[h!]
\small
\caption{A comprehensive performance analysis of our various proposed frameworks.}
\centering
\label{tab: acc_study}
\begin{tabular}{cc|c|c|c}
\hline
\multicolumn{2}{c|}{\textbf{Framework}}                                                                           & \textbf{CNN} & \textbf{CRNN} & \textbf{\begin{tabular}[c]{@{}c@{}}CRNN \\ with \\ Attention\end{tabular}} \\ \hline\hline
\multicolumn{2}{c|}{\textbf{Parameters}}                                                                           & 1,355,917 & 2,094,477 & 2,357,645 \\ \hline\hline
\multicolumn{2}{c|}{\textbf{Indian Dataset}}                                                    & 0.983        & \textbf{0.987}         & \textbf{0.987}                                                                      \\ \hline
\multicolumn{1}{c|}{\multirow{3}{*}{\textbf{Close Language Cluster}}}    & \textbf{Cluster 1}   & 0.971        & \text{0.974}         & \textbf{0.980}                                                                      \\ \cline{2-5} 
\multicolumn{1}{c|}{}                                                    & \textbf{Cluster 2}   & 0.996        & \textbf{0.999}         & \textbf{0.999}                                                                      \\ \cline{2-5} 
\multicolumn{1}{c|}{}                                                    & \textbf{Cluster 3}   & 0.996        & \textbf{1}             & \text{0.999}                                                                      \\ \hline
\multicolumn{1}{c|}{\multirow{2}{*}{\textbf{European Language Dataset}}} & \textbf{No Noise}    & 0.948        & \textbf{0.967}         & \text{0.966}                                                                      \\ \cline{2-5} 
\multicolumn{1}{c|}{}                                                    & \textbf{White Noise} & 0.871        & \textbf{0.912}         & \text{0.888}                                                                      \\ \hline
\end{tabular}
\end{table}
Table \ref{tab: acc_study} demonstrates that both CRNN and CRNN with attention perform better compared to the CNN-based framework. At the same time, CRNN itself produces better or equivalent performance compared to CRNN with an Attention-based mechanism. CRNN with Attention performs better only for Cluster 1 of the Indian dataset; CRNN itself produces the best results in all other tasks, sometimes jointly with CRNN with Attention. This is despite the fact that the Attention-based framework has more parameters than the other models. The underlying intuition is that the attention-based framework generally suffers from overfitting problems due to its additional parameter count. An attention-based framework needs to learn how to assign importance to different parts of the input sequence, which may require a large number of training instances to produce a generalized performance. Thus, CRNN with Attention makes the experimental set-up time-consuming and resource-intensive, but still, it is not able to improve over CRNN.

\section{Conclusion and future work}
In this work, we proposed a language identification method using CRNN that works on MFCC features of speech signals. Our framework efficiently identifies the language both in close language and noisy scenarios. We carried out extensive experiments and our framework produced state-of-the-art results. Through our experiments, we have also shown our framework's robustness to noise and its extensibility to new languages. The model exhibits the overall best accuracy of 98.7\% which improves over the traditional use of CNN (98.3\%). CRNN with attention performs almost at par with CRNN, however, the attention mechanism which incurs additional computational overhead does not result in improvement over CRNN in most cases. \par
In future, we would like to extend our work by increasing the language classes with speech specimens recorded in different environments. We would also like to extend our work to check the usefulness of the proposed framework on smaller time speech samples through which we can deduce the optimal time required to classify the languages with high accuracy. We would also like to test our method on language dialect identification.

\section*{Acknowledgements}
This research was supported by the TPU Research Cloud (TRC) program, a Google Research initiative and funded by Rashtriya Uchchatar Shiksha Abhiyan 2.0 [grant number R-11/828/19].

\newpage
\appendix

\section{CNN framework}
\setcounter{table}{0}
\renewcommand{\thetable}{A.\arabic{table}}

\begin{table}[h!]
\tiny
\caption{Confusion matrix of Manually Balancing the Samples for each category to 100 with CNN}
\centering
\label{tab: 13_category_100_cnn}
	\begin{tabular}{c|c|c|c|c|c|c|c|c|c|c|c|c|c|c|c|c|c }
    \hline
    \multicolumn{2}{c|}{\multirow{2}{*}{}} & \multicolumn{13}{c|}{\textbf{Predicted}} & \multirow{2}{*}{\textbf{PPV}} & \multirow{2}{*}{\textbf{TPR}} & \multirow{2}{*}{\textbf{\begin{tabular}[c]{@{}c@{}}f1\\[-0.5em] Score\end{tabular}}} \\ \cline{3-15}
    \multicolumn{2}{c|}{} & \textbf{as} & \textbf{bn} & \textbf{bd} & \textbf{gu} & \textbf{hi} & \textbf{kn} & \textbf{ml} & \textbf{mn} & \textbf{mr} & \textbf{or} & \textbf{rj} & \textbf{ta} & \textbf{te} &  &  &  \\
    \hline
    \parbox[t]{2mm}{\multirow{13}{*}{\rotatebox[origin=c]{90}{\textbf{Actual}}}} & \textbf{as} & \textbf{29} & 2 & 0 & 1 & 0 & 3 & 0 & 7 & 0 & 0 & 3 & 2 & 3 & 0.617 & 0.58 & 0.598 \\ \cline{2-18} 
    & \textbf{bn} & 2 & \textbf{40} & 0 & 0 & 0 & 1 & 1 & 1 & 1 & 0 & 0 & 3 & 1 & 0.816 & 0.8 & 0.808 \\ \cline{2-18} 
    & \textbf{bd} & 0 & 0 & \textbf{43} & 1 & 0 & 2 & 0 & 0 & 2 & 0 & 1 & 1 & 0 & 0.843 & 0.86 & 0.851 \\ \cline{2-18} 
    & \textbf{gu} & 1 & 0 & 0 & \textbf{38} & 4 & 0 & 0 & 0 & 0 & 0 & 0 & 3 & 4 & 0.731 & 0.76 & 0.745 \\ \cline{2-18} 
    & \textbf{hi} & 2 & 0 & 0 & 2 & \textbf{35} & 3 & 3 & 0 & 0 & 2 & 0 & 2 & 1 & 0.778 & 0.7 & 0.737 \\ \cline{2-18} 
    & \textbf{kn} & 0 & 0 & 1 & 0 & 1 & \textbf{37} & 6 & 1 & 0 & 0 & 1 & 0 & 3 & 0.725 & 0.74 & 0.733 \\ \cline{2-18} 
    & \textbf{ml} & 0 & 1 & 0 & 1 & 0 & 3 & \textbf{41} & 0 & 0 & 0 & 2 & 1 & 1 & 0.774 & 0.82 & 0.796 \\ \cline{2-18} 
    & \textbf{mn} & 2 & 2 & 0 & 0 & 0 & 0 & 1 & \textbf{38} & 0 & 1 & 0 & 1 & 5 & 0.691 & 0.76 & 0.724 \\ \cline{2-18} 
    & \textbf{mr} & 0 & 0 & 1 & 0 & 0 & 0 & 0 & 0 & \textbf{42} & 3 & 4 & 0 & 0 & 0.857 & 0.84 & 0.848 \\ \cline{2-18} 
    & \textbf{or} & 0 & 1 & 1 & 0 & 0 & 0 & 0 & 0 & 3 & \textbf{43} & 1 & 1 & 0 & 0.811 & 0.86 & 0.835 \\ \cline{2-18} 
    & \textbf{rj} & 2 & 2 & 3 & 1 & 3 & 0 & 1 & 2 & 1 & 2 & \textbf{26} & 7 & 0 & 0.605 & 0.52 & 0.559 \\ \cline{2-18} 
    & \textbf{ta} & 5 & 0 & 0 & 3 & 2 & 0 & 0 & 3 & 0 & 0 & 2 & \textbf{31} & 4 & 0.564 & 0.62 & 0.590 \\ \cline{2-18} 
    & \textbf{te} & 4 & 1 & 2 & 5 & 0 & 2 & 0 & 3 & 0 & 2 & 3 & 3 & \textbf{25} & 0.532 & 0.5 & 0.515 \\
    \hline
    \end{tabular}
\end{table}

\begin{table}[h!]
\tiny
\caption{Confusion matrix of Manually Balancing the Samples for each category to 200 with CNN} 
\centering
\label{tab: 13_category_200_cnn}
\begin{tabular}{c|c|c|c|c|c|c|c|c|c|c|c|c|c|c|c|c|c }
    \hline
    \multicolumn{2}{c|}{\multirow{2}{*}{}} & \multicolumn{13}{c|}{\textbf{Predicted}} & \multirow{2}{*}{\textbf{PPV}} & \multirow{2}{*}{\textbf{TPR}} & \multirow{2}{*}{\textbf{\begin{tabular}[c]{@{}c@{}}f1\\[-0.5em] Score\end{tabular}}} \\ \cline{3-15}
    \multicolumn{2}{c|}{} & \textbf{as} & \textbf{bn} & \textbf{bd} & \textbf{gu} & \textbf{hi} & \textbf{kn} & \textbf{ml} & \textbf{mn} & \textbf{mr} & \textbf{or} & \textbf{rj} & \textbf{ta} & \textbf{te} &  &  &  \\
    \hline
    \parbox[t]{2mm}{\multirow{13}{*}{\rotatebox[origin=c]{90}{\textbf{Actual}}}} & \textbf{as} & \textbf{44} & 1 & 0 & 0 & 0 & 0 & 0 & 1 & 0 & 0 & 1 & 3 & 0 & 0.8 & 0.88 & 0.838 \\ \cline{2-18} 
 & \textbf{bn} & 1 & \textbf{46} & 0 & 0 & 0 & 0 & 0 & 1 & 0 & 0 & 0 & 2 & 0 & 0.92 & 0.92 & 0.92 \\ \cline{2-18} 
 & \textbf{bd} & 0 & 0 & \textbf{47} & 0 & 0 & 1 & 0 & 1 & 0 & 0 & 1 & 0 & 0 & 0.94 & 0.94 & 0.94 \\ \cline{2-18} 
 & \textbf{gu} & 1 & 1 & 0 & \textbf{45} & 1 & 0 & 0 & 0 & 0 & 0 & 0 & 0 & 2 & 0.918 & 0.9 & 0.909 \\ \cline{2-18} 
 & \textbf{hi} & 0 & 1 & 0 & 1 & \textbf{43} & 1 & 1 & 2 & 0 & 0 & 1 & 0 & 0 & 0.956 & 0.86 & 0.905 \\ \cline{2-18} 
 & \textbf{kn} & 0 & 0 & 1 & 1 & 0 & \textbf{43} & 2 & 1 & 0 & 0 & 0 & 1 & 1 & 0.878 & 0.86 & 0.869 \\ \cline{2-18} 
 & \textbf{ml} & 1 & 0 & 0 & 0 & 0 & 1 & \textbf{43} & 2 & 1 & 2 & 0 & 0 & 0 & 0.896 & 0.86 & 0.878 \\ \cline{2-18} 
 & \textbf{mn} & 2 & 0 & 0 & 0 & 0 & 0 & 1 & \textbf{46} & 0 & 0 & 0 & 1 & 0 & 0.754 & 0.92 & 0.829 \\ \cline{2-18} 
 & \textbf{mr} & 0 & 0 & 2 & 0 & 0 & 0 & 1 & 1 & \textbf{43} & 1 & 2 & 0 & 0 & 0.956 & 0.86 & 0.905 \\ \cline{2-18} 
 & \textbf{or} & 0 & 0 & 0 & 0 & 1 & 0 & 0 & 0 & 0 & \textbf{48} & 1 & 0 & 0 & 0.941 & 0.96 & 0.95 \\ \cline{2-18} 
 & \textbf{rj} & 2 & 1 & 0 & 0 & 0 & 1 & 0 & 1 & 1 & 0 & \textbf{43} & 1 & 0 & 0.86 & 0.86 & 0.86 \\ \cline{2-18} 
 & \textbf{ta} & 2 & 0 & 0 & 1 & 0 & 2 & 0 & 4 & 0 & 0 & 1 & \textbf{40} & 0 & 0.784 & 0.8 & 0.792 \\ \cline{2-18} 
 & \textbf{te} & 2 & 0 & 0 & 1 & 0 & 0 & 0 & 1 & 0 & 0 & 0 & 3 & \textbf{43} & 0.935 & 0.86 & 0.896  \\
    \hline
    \end{tabular}
\end{table}

\begin{table}[h!]
\tiny
\caption{Confusion matrix of Manually Balancing the Samples for each category to 571 with CNN}
\centering
\label{tab: 13_category_571_cnn}
 \begin{tabular}{ c|c|c|c|c|c|c|c|c|c|c|c|c|c|c|c|c|c}
    \hline
    \multicolumn{2}{c|}{\multirow{2}{*}{}} & \multicolumn{13}{c|}{\textbf{Predicted}} & \multirow{2}{*}{\textbf{PPV}} & \multirow{2}{*}{\textbf{TPR}} & \multirow{2}{*}{\textbf{\begin{tabular}[c]{@{}c@{}}f1\\[-0.5em] Score\end{tabular}}} \\ \cline{3-15}
\multicolumn{2}{c|}{} & \textbf{as} & \textbf{bn} & \textbf{bd} & \textbf{gu} & \textbf{hi} & \textbf{kn} & \textbf{ml} & \textbf{mn} & \textbf{mr} & \textbf{or} & \textbf{rj} & \textbf{ta} & \textbf{te} &  &  &  \\
    \hline
    \parbox[t]{2mm}{\multirow{13}{*}{\rotatebox[origin=c]{90}{\textbf{Actual}}}} & \textbf{as} & \textbf{58} & 0 & 0 & 0 & 0 & 0 & 0 & 0 & 0 & 0 & 0 & 0 & 0 & 0.967 & 1 & 0.983 \\ \cline{2-18} 
 & \textbf{bn} & 0 & \textbf{58} & 0 & 0 & 0 & 0 & 0 & 0 & 0 & 0 & 0 & 0 & 0 & 0.983 & 1 & 0.991 \\ \cline{2-18} 
 & \textbf{bd} & 0 & 0 & \textbf{56} & 0 & 0 & 0 & 0 & 0 & 0 & 0 & 0 & 0 & 0 & 1 & 1 & 1 \\ \cline{2-18} 
 & \textbf{gu} & 0 & 0 & 0 & \textbf{54} & 4 & 0 & 0 & 0 & 0 & 0 & 0 & 0 & 0 & 0.982 & 0.931 & 0.956 \\ \cline{2-18} 
 & \textbf{hi} & 0 & 0 & 0 & 0 & \textbf{50} & 0 & 2 & 1 & 0 & 0 & 0 & 5 & 0 & 0.893 & 0.862 & 0.877 \\ \cline{2-18} 
 & \textbf{kn} & 0 & 0 & 0 & 0 & 0 & \textbf{56} & 2 & 0 & 0 & 0 & 0 & 0 & 0 & 0.903 & 0.966 & 0.933 \\ \cline{2-18} 
 & \textbf{ml} & 0 & 0 & 0 & 0 & 0 & 4 & \textbf{53} & 0 & 0 & 0 & 1 & 0 & 0 & 0.914 & 0.914 & 0.914 \\ \cline{2-18} 
 & \textbf{mn} & 1 & 0 & 0 & 0 & 0 & 1 & 0 & \textbf{54} & 0 & 0 & 0 & 1 & 1 & 0.931 & 0.931 & 0.931 \\ \cline{2-18} 
 & \textbf{mr} & 0 & 0 & 0 & 0 & 0 & 0 & 0 & 0 & \textbf{55} & 0 & 1 & 0 & 0 & 0.965 & 0.982 & 0.973 \\ \cline{2-18} 
 & \textbf{or} & 0 & 0 & 0 & 0 & 1 & 0 & 0 & 0 & 1 & \textbf{56} & 0 & 0 & 0 & 1 & 0.966 & 0.982 \\ \cline{2-18} 
 & \textbf{rj} & 1 & 1 & 0 & 0 & 0 & 0 & 1 & 1 & 0 & 0 & \textbf{54} & 0 & 0 & 0.9 & 0.931 & 0.915 \\ \cline{2-18} 
 & \textbf{ta} & 0 & 0 & 0 & 1 & 1 & 0 & 0 & 1 & 1 & 0 & 3 & \textbf{51} & 0 & 0.879 & 0.879 & 0.879 \\ \cline{2-18} 
 & \textbf{te} & 0 & 0 & 0 & 0 & 0 & 1 & 0 & 1 & 0 & 0 & 1 & 1 & \textbf{54} & 0.982 & 0.931 & 0.956 \\
    \hline
    \end{tabular}
\end{table}

\section{CRNN framework}
\setcounter{table}{0}
\renewcommand{\thetable}{B.\arabic{table}}

\begin{table}[h!]
\tiny
\caption{Confusion matrix of Manually Balancing the Samples for each category to 100 with CRNN}
\centering
\label{tab: 13_category_100_CRNN}
  \begin{tabular}{c|c|c|c|c|c|c|c|c|c|c|c|c|c|c|c|c|c }
    \hline
    \multicolumn{2}{c|}{\multirow{2}{*}{}} & \multicolumn{13}{c|}{\textbf{Predicted}} & \multirow{2}{*}{\textbf{PPV}} & \multirow{2}{*}{\textbf{TPR}} & \multirow{2}{*}{\textbf{\begin{tabular}[c]{@{}c@{}}f1\\[-0.5em] Score\end{tabular}}} \\ \cline{3-15}
    \multicolumn{2}{c|}{} & \textbf{as} & \textbf{bn} & \textbf{bd} & \textbf{gu} & \textbf{hi} & \textbf{kn} & \textbf{ml} & \textbf{mn} & \textbf{mr} & \textbf{or} & \textbf{rj} & \textbf{ta} & \textbf{te} &  &  &  \\
    \hline
    \parbox[t]{2mm}{\multirow{13}{*}{\rotatebox[origin=c]{90}{\textbf{Actual}}}} & \textbf{as} & \textbf{47} & 0 & 0 & 0 & 0 & 0 & 0 & 1 & 0 & 0 & 1 & 1 & 0 & 0.839 & 0.94 & 0.887 \\ \cline{2-18} 
 & \textbf{bn} & 0 & \textbf{45} & 0 & 0 & 0 & 1 & 0 & 0 & 0 & 0 & 0 & 3 & 1 & 0.957 & 0.9 & 0.928 \\ \cline{2-18} 
 & \textbf{bd} & 0 & 0 & \textbf{50} & 0 & 0 & 0 & 0 & 0 & 0 & 0 & 0 & 0 & 0 & 0.962 & 1 & 0.98 \\ \cline{2-18} 
 & \textbf{gu} & 0 & 0 & 0 & \textbf{49} & 0 & 0 & 0 & 0 & 0 & 0 & 0 & 0 & 1 & 1 & 0.98 & 0.99 \\ \cline{2-18} 
 & \textbf{hi} & 0 & 0 & 0 & 0 & \textbf{45} & 2 & 2 & 0 & 0 & 0 & 0 & 0 & 1 & 0.957 & 0.9 & 0.928 \\ \cline{2-18} 
 & \textbf{kn} & 1 & 0 & 0 & 0 & 0 & \textbf{47} & 2 & 0 & 0 & 0 & 0 & 0 & 0 & 0.94 & 0.94 & 0.94 \\ \cline{2-18} 
 & \textbf{ml} & 0 & 0 & 0 & 0 & 1 & 0 & \textbf{48} & 0 & 0 & 0 & 1 & 0 & 0 & 0.923 & 0.96 & 0.941 \\ \cline{2-18} 
 & \textbf{mn} & 1 & 2 & 0 & 0 & 0 & 0 & 0 & \textbf{43} & 0 & 1 & 1 & 0 & 2 & 0.935 & 0.86 & 0.896 \\ \cline{2-18} 
 & \textbf{mr} & 0 & 0 & 0 & 0 & 0 & 0 & 0 & 0 & \textbf{48} & 0 & 2 & 0 & 0 & 0.98 & 0.96 & 0.97 \\ \cline{2-18} 
 & \textbf{or} & 0 & 0 & 0 & 0 & 0 & 0 & 0 & 0 & 0 & \textbf{50} & 0 & 0 & 0 & 0.943 & 1 & 0.971 \\ \cline{2-18} 
 & \textbf{rj} & 4 & 0 & 1 & 0 & 0 & 0 & 0 & 0 & 1 & 1 & \textbf{42} & 1 & 0 & 0.894 & 0.84 & 0.866 \\ \cline{2-18} 
 & \textbf{ta} & 2 & 0 & 0 & 0 & 1 & 0 & 0 & 2 & 0 & 1 & 0 & \textbf{44} & 0 & 0.898 & 0.88 & 0.889 \\ \cline{2-18} 
 & \textbf{te} & 1 & 0 & 1 & 0 & 0 & 0 & 0 & 0 & 0 & 0 & 0 & 0 & \textbf{48} & 0.906 & 0.96 & 0.932 \\
    \hline
    \end{tabular}
\end{table}

\begin{table}[h!]
\tiny
\caption{Confusion matrix of Manually Balancing the Samples for each category to 200 with CRNN}
\centering
\label{tab: 13_category_200_CRNN}
 \begin{tabular}{ c|c|c|c|c|c|c|c|c|c|c|c|c|c|c|c|c|c }
    \hline
  \multicolumn{2}{c|}{\multirow{2}{*}{}} & \multicolumn{13}{c|}{\textbf{Predicted}} & \multirow{2}{*}{\textbf{PPV}} & \multirow{2}{*}{\textbf{TPR}} & \multirow{2}{*}{\textbf{\begin{tabular}[c]{@{}c@{}}f1\\[-0.5em] Score\end{tabular}}} \\ \cline{3-15}
  \multicolumn{2}{c|}{} & \textbf{as} & \textbf{bn} & \textbf{bd} & \textbf{gu} & \textbf{hi} & \textbf{kn} & \textbf{ml} & \textbf{mn} & \textbf{mr} & \textbf{or} & \textbf{rj} & \textbf{ta} & \textbf{te} &  &  &  \\
    \hline
  \parbox[t]{2mm}{\multirow{13}{*}{\rotatebox[origin=c]{90}{\textbf{Actual}}}} & \textbf{as} & \textbf{47} & 0 & 0 & 0 & 0 & 0 & 0 & 2 & 0 & 0 & 0 & 1 & 0 & 1 & 0.94 & 0.969 \\ \cline{2-18} 
  & \textbf{bn} & 0 & \textbf{48} & 0 & 0 & 0 & 0 & 0 & 0 & 0 & 0 & 0 & 2 & 0 & 1 & 0.96 & 0.98 \\ \cline{2-18} 
  & \textbf{bd} & 0 & 0 & \textbf{49} & 0 & 0 & 0 & 1 & 0 & 0 & 0 & 0 & 0 & 0 & 0.98 & 0.98 & 0.98 \\ \cline{2-18} 
  & \textbf{gu} & 0 & 0 & 0 & \textbf{50} & 0 & 0 & 0 & 0 & 0 & 0 & 0 & 0 & 0 & 1 & 1 & 1 \\ \cline{2-18} 
  & \textbf{hi} & 0 & 0 & 0 & 0 & \textbf{49} & 0 & 1 & 0 & 0 & 0 & 0 & 0 & 0 & 1 & 0.98 & 0.99 \\ \cline{2-18} 
  & \textbf{kn} & 0 & 0 & 0 & 0 & 0 & \textbf{49} & 1 & 0 & 0 & 0 & 0 & 0 & 0 & 1 & 0.98 & 0.99 \\ \cline{2-18} 
  & \textbf{ml} & 0 & 0 & 0 & 0 & 0 & 0 & \textbf{50} & 0 & 0 & 0 & 0 & 0 & 0 & 0.893 & 1 & 0.943 \\ \cline{2-18} 
  & \textbf{mn} & 0 & 0 & 0 & 0 & 0 & 0 & 1 & \textbf{49} & 0 & 0 & 0 & 0 & 0 & 0.907 & 0.98 & 0.942 \\ \cline{2-18} 
  & \textbf{mr} & 0 & 0 & 1 & 0 & 0 & 0 & 0 & 1 & \textbf{48} & 0 & 0 & 0 & 0 & 0.98 & 0.96 & 0.97 \\ \cline{2-18} 
  & \textbf{or} & 0 & 0 & 0 & 0 & 0 & 0 & 0 & 0 & 0 & \textbf{50} & 0 & 0 & 0 & 1 & 1 & 1 \\ \cline{2-18} 
  & \textbf{rj} & 0 & 0 & 0 & 0 & 0 & 0 & 0 & 0 & 1 & 0 & \textbf{48} & 1 & 0 & 1 & 0.96 & 0.98 \\ \cline{2-18} 
  & \textbf{ta} & 0 & 0 & 0 & 0 & 0 & 0 & 0 & 2 & 0 & 0 & 0 & \textbf{47} & 1 & 0.904 & 0.94 & 0.922 \\ \cline{2-18} 
  & \textbf{te} & 0 & 0 & 0 & 0 & 0 & 0 & 2 & 0 & 0 & 0 & 0 & 1 & \textbf{47} & 0.979 & 0.94 & 0.959 \\
    \hline
    \end{tabular}
\end{table}

\begin{table}[h!]
\tiny
\caption{Confusion matrix of Manually Balancing the Samples for each category to 571 with CRNN}
\centering
\label{tab: 13_category_571_CRNN}
\begin{tabular}{ c|c|c|c|c|c|c|c|c|c|c|c|c|c|c|c|c|c }
    \hline
  \multicolumn{2}{c|}{\multirow{2}{*}{}} & \multicolumn{13}{c|}{\textbf{Predicted}} & \multirow{2}{*}{\textbf{PPV}} & \multirow{2}{*}{\textbf{TPR}} & \multirow{2}{*}{\textbf{\begin{tabular}[c]{@{}c@{}}f1\\[-0.5em] Score\end{tabular}}} \\ \cline{3-15}
  \multicolumn{2}{c|}{} & \textbf{as} & \textbf{bn} & \textbf{bd} & \textbf{gu} & \textbf{hi} & \textbf{kn} & \textbf{ml} & \textbf{mn} & \textbf{mr} & \textbf{or} & \textbf{rj} & \textbf{ta} & \textbf{te} &  &  &  \\
    \hline
  \parbox[t]{2mm}{\multirow{13}{*}{\rotatebox[origin=c]{90}{\textbf{Actual}}}} & \textbf{as} & \textbf{57} & 0 & 0 & 0 & 0 & 0 & 0 & 0 & 0 & 0 & 0 & 1 & 0 & 0.983 & 0.983 & 0.983 \\ \cline{2-18} 
  & \textbf{bn} & 0 & \textbf{58} & 0 & 0 & 0 & 0 & 0 & 0 & 0 & 0 & 0 & 0 & 0 & 1 & 1 & 1 \\ \cline{2-18} 
  & \textbf{bd} & 0 & 0 & \textbf{56} & 0 & 0 & 0 & 0 & 0 & 0 & 0 & 0 & 0 & 0 & 1 & 1 & 1 \\ \cline{2-18} 
  & \textbf{gu} & 0 & 0 & 0 & \textbf{58} & 0 & 0 & 0 & 0 & 0 & 0 & 0 & 0 & 0 & 0.983 & 1 & 0.991 \\ \cline{2-18} 
  & \textbf{hi} & 0 & 0 & 0 & 0 & \textbf{58} & 0 & 0 & 0 & 0 & 0 & 0 & 0 & 0 & 1 & 1 & 1 \\ \cline{2-18} 
  & \textbf{kn} & 0 & 0 & 0 & 0 & 0 & \textbf{58} & 0 & 0 & 0 & 0 & 0 & 0 & 0 & 1 & 1 & 1 \\ \cline{2-18} 
  & \textbf{ml} & 0 & 0 & 0 & 0 & 0 & 0 & \textbf{58} & 0 & 0 & 0 & 0 & 0 & 0 & 0.983 & 1 & 0.991 \\ \cline{2-18} 
  & \textbf{mn} & 0 & 0 & 0 & 0 & 0 & 0 & 0 & \textbf{57} & 0 & 0 & 1 & 0 & 0 & 1 & 0.983 & 0.991 \\ \cline{2-18} 
  & \textbf{mr} & 0 & 0 & 0 & 0 & 0 & 0 & 0 & 0 & \textbf{56} & 0 & 0 & 0 & 0 & 0.982 & 1 & 0.991 \\ \cline{2-18} 
  & \textbf{or} & 0 & 0 & 0 & 0 & 0 & 0 & 0 & 0 & 1 & \textbf{57} & 0 & 0 & 0 & 1 & 0.983 & 0.991 \\ \cline{2-18} 
  & \textbf{rj} & 1 & 0 & 0 & 0 & 0 & 0 & 1 & 0 & 0 & 0 & \textbf{56} & 0 & 0 & 0.918 & 0.966 & 0.941 \\ \cline{2-18} 
  & \textbf{ta} & 0 & 0 & 0 & 1 & 0 & 0 & 0 & 0 & 0 & 0 & 4 & \textbf{53} & 0 & 0.964 & 0.914 & 0.938 \\ \cline{2-18} 
  & \textbf{te} & 0 & 0 & 0 & 0 & 0 & 0 & 0 & 0 & 0 & 0 & 0 & 1 & \textbf{57} & 1 & 0.983 & 0.991 \\
    \hline
    \end{tabular}
\end{table}

\section{CRNN with Attention framework}
\setcounter{table}{0}
\renewcommand{\thetable}{C.\arabic{table}}

\begin{table}[h!]
\tiny
\caption{Confusion matrix of Manually Balancing the Samples for each category to 100 with CRNN and Attention}
\centering
\label{tab: 13_category_100_att}
\begin{tabular}{c|c|c|c|c|c|c|c|c|c|c|c|c|c|c|c|c|c}
    \hline
  \multicolumn{2}{c|}{\multirow{2}{*}{}} & \multicolumn{13}{c|}{\textbf{Predicted}} & \multirow{2}{*}{\textbf{PPV}} & \multirow{2}{*}{\textbf{TPR}} & \multirow{2}{*}{\textbf{\begin{tabular}[c]{@{}c@{}}f1\\[-0.5em] Score\end{tabular}}} \\ \cline{3-15}
  \multicolumn{2}{c|}{} & \textbf{as} & \textbf{bn} & \textbf{bd} & \textbf{gu} & \textbf{hi} & \textbf{kn} & \textbf{ml} & \textbf{mn} & \textbf{mr} & \textbf{or} & \textbf{rj} & \textbf{ta} & \textbf{te} &  &  &  \\
    \hline
  \parbox[t]{2mm}{\multirow{13}{*}{\rotatebox[origin=c]{90}{\textbf{Actual}}}} & \textbf{as} & \textbf{36} & 0 & 0 & 0 & 0 & 0 & 0 & 1 & 0 & 0 & 6 & 6 & 1 & 0.766 & 0.72 & 0.742 \\ \cline{2-18} 
  & \textbf{bn} & 1 & \textbf{35} & 0 & 0 & 0 & 0 & 0 & 1 & 0 & 0 & 0 & 13 & 0 & 0.875 & 0.7 & 0.778 \\ \cline{2-18} 
  & \textbf{bd} & 0 & 0 & \textbf{50} & 0 & 0 & 0 & 0 & 0 & 0 & 0 & 0 & 0 & 0 & 1 & 1 & 1 \\ \cline{2-18} 
  & \textbf{gu} & 0 & 0 & 0 & \textbf{50} & 0 & 0 & 0 & 0 & 0 & 0 & 0 & 0 & 0 & 0.943 & 1 & 0.971 \\ \cline{2-18} 
  & \textbf{hi} & 0 & 0 & 0 & 0 & \textbf{47} & 1 & 1 & 0 & 0 & 0 & 0 & 0 & 1 & 0.959 & 0.94 & 0.95 \\ \cline{2-18} 
  & \textbf{kn} & 0 & 0 & 0 & 0 & 0 & \textbf{49} & 0 & 0 & 0 & 0 & 0 & 1 & 0 & 0.961 & 0.98 & 0.97 \\ \cline{2-18} 
  & \textbf{ml} & 0 & 0 & 0 & 0 & 1 & 1 & \textbf{46} & 0 & 0 & 0 & 2 & 0 & 0 & 0.958 & 0.92 & 0.939 \\ \cline{2-18} 
  & \textbf{mn} & 5 & 3 & 0 & 1 & 0 & 0 & 0 & \textbf{36} & 0 & 1 & 1 & 0 & 3 & 0.878 & 0.72 & 0.791 \\ \cline{2-18} 
  & \textbf{mr} & 0 & 0 & 0 & 0 & 0 & 0 & 0 & 0 & \textbf{48} & 0 & 2 & 0 & 0 & 0.906 & 0.96 & 0.932 \\ \cline{2-18} 
  & \textbf{or} & 0 & 0 & 0 & 0 & 0 & 0 & 0 & 0 & 3 & \textbf{47} & 0 & 0 & 0 & 0.959 & 0.94 & 0.949 \\ \cline{2-18} 
  & \textbf{rj} & 2 & 0 & 0 & 0 & 0 & 0 & 1 & 0 & 2 & 1 & \textbf{43} & 1 & 0 & 0.782 & 0.86 & 0.819 \\ \cline{2-18} 
  & \textbf{ta} & 2 & 1 & 0 & 0 & 1 & 0 & 0 & 0 & 0 & 0 & 1 & \textbf{44} & 1 & 0.677 & 0.88 & 0.765 \\ \cline{2-18} 
  & \textbf{te} & 1 & 1 & 0 & 2 & 0 & 0 & 0 & 3 & 0 & 0 & 0 & 0 & \textbf{43} & 0.878 & 0.86 & 0.869 \\
    \hline
    \end{tabular}
\end{table}

\begin{table}[h!]
\tiny
\caption{Confusion matrix of Manually Balancing the Samples for each category to 200 with CRNN and Attention}
\centering
\label{tab: 13_category_200_att}
 \begin{tabular}{ c|c|c|c|c|c|c|c|c|c|c|c|c|c|c|c|c|c }
    \hline
  \multicolumn{2}{c|}{\multirow{2}{*}{}} & \multicolumn{13}{c|}{\textbf{Predicted}} & \multirow{2}{*}{\textbf{PPV}} & \multirow{2}{*}{\textbf{TPR}} & \multirow{2}{*}{\textbf{\begin{tabular}[c]{@{}c@{}}f1\\[-0.5em] Score\end{tabular}}} \\ \cline{3-15}
  \multicolumn{2}{c|}{} & \textbf{as} & \textbf{bn} & \textbf{bd} & \textbf{gu} & \textbf{hi} & \textbf{kn} & \textbf{ml} & \textbf{mn} & \textbf{mr} & \textbf{or} & \textbf{rj} & \textbf{ta} & \textbf{te} &  &  &  \\
    \hline
  \parbox[t]{2mm}{\multirow{13}{*}{\rotatebox[origin=c]{90}{\textbf{Actual}}}} & \textbf{as} & \textbf{48} & 2 & 0 & 0 & 0 & 0 & 0 & 0 & 0 & 0 & 0 & 0 & 0 & 0.941 & 0.96 & 0.95 \\ \cline{2-18} 
 & \textbf{bn} & 0 & \textbf{50} & 0 & 0 & 0 & 0 & 0 & 0 & 0 & 0 & 0 & 0 & 0 & 0.909 & 1 & 0.952 \\ \cline{2-18} 
 & \textbf{bd} & 0 & 0 & \textbf{48} & 0 & 0 & 0 & 0 & 0 & 1 & 0 & 1 & 0 & 0 & 0.98 & 0.96 & 0.97 \\ \cline{2-18} 
 & \textbf{gu} & 0 & 0 & 0 & \textbf{50} & 0 & 0 & 0 & 0 & 0 & 0 & 0 & 0 & 0 & 1 & 1 & 1 \\ \cline{2-18} 
 & \textbf{hi} & 0 & 0 & 0 & 0 & \textbf{49} & 0 & 1 & 0 & 0 & 0 & 0 & 0 & 0 & 1 & 0.98 & 0.99 \\ \cline{2-18} 
 & \textbf{kn} & 0 & 0 & 0 & 0 & 0 & \textbf{49} & 1 & 0 & 0 & 0 & 0 & 0 & 0 & 1 & 0.98 & 0.99 \\ \cline{2-18} 
 & \textbf{ml} & 0 & 0 & 0 & 0 & 0 & 0 & \textbf{50} & 0 & 0 & 0 & 0 & 0 & 0 & 0.962 & 1 & 0.98 \\ \cline{2-18} 
 & \textbf{mn} & 3 & 0 & 0 & 0 & 0 & 0 & 0 & \textbf{46} & 0 & 1 & 0 & 0 & 0 & 0.979 & 0.92 & 0.948 \\ \cline{2-18} 
 & \textbf{mr} & 0 & 0 & 1 & 0 & 0 & 0 & 0 & 0 & \textbf{49} & 0 & 0 & 0 & 0 & 0.98 & 0.98 & 0.98 \\ \cline{2-18} 
 & \textbf{or} & 0 & 0 & 0 & 0 & 0 & 0 & 0 & 0 & 0 & \textbf{50} & 0 & 0 & 0 & 0.980 & 1 & 0.99 \\ \cline{2-18} 
 & \textbf{rj} & 0 & 2 & 0 & 0 & 0 & 0 & 0 & 0 & 0 & 0 & \textbf{48} & 0 & 0 & 0.96 & 0.96 & 0.96 \\ \cline{2-18} 
 & \textbf{ta} & 0 & 1 & 0 & 0 & 0 & 0 & 0 & 0 & 0 & 0 & 1 & \textbf{48} & 0 & 1 & 0.96 & 0.98 \\ \cline{2-18} 
 & \textbf{te} & 0 & 0 & 0 & 0 & 0 & 0 & 0 & 1 & 0 & 0 & 0 & 0 & \textbf{49} & 1 & 0.98 & 0.99 \\
    \hline
    \end{tabular}
\end{table}

\begin{table}[h!]
\tiny
\caption{Confusion matrix of Manually Balancing the Samples for each category to 571 with CRNN and Attention}
\centering
\label{tab: 13_category_571_att}
\begin{tabular}{ c|c|c|c|c|c|c|c|c|c|c|c|c|c|c|c|c|c }
    \hline
  \multicolumn{2}{c|}{\multirow{2}{*}{}} & \multicolumn{13}{c|}{\textbf{Predicted}} & \multirow{2}{*}{\textbf{PPV}} & \multirow{2}{*}{\textbf{TPR}} & \multirow{2}{*}{\textbf{\begin{tabular}[c]{@{}c@{}}f1\\[-0.5em] Score\end{tabular}}} \\ \cline{3-15}
  \multicolumn{2}{c|}{} & \textbf{as} & \textbf{bn} & \textbf{bd} & \textbf{gu} & \textbf{hi} & \textbf{kn} & \textbf{ml} & \textbf{mn} & \textbf{mr} & \textbf{or} & \textbf{rj} & \textbf{ta} & \textbf{te} &  &  &  \\
    \hline
  \parbox[t]{2mm}{\multirow{13}{*}{\rotatebox[origin=c]{90}{\textbf{Actual}}}} & \textbf{as} & \textbf{58} & 0 & 0 & 0 & 0 & 0 & 0 & 0 & 0 & 0 & 0 & 0 & 0 & 1 & 1 & 1 \\ \cline{2-18} 
 & \textbf{bn} & 0 & \textbf{58} & 0 & 0 & 0 & 0 & 0 & 0 & 0 & 0 & 0 & 0 & 0 & 1 & 1 & 1 \\ \cline{2-18} 
 & \textbf{bd} & 0 & 0 & \textbf{56} & 0 & 0 & 0 & 0 & 0 & 0 & 0 & 0 & 0 & 0 & 1 & 1 & 1 \\ \cline{2-18} 
 & \textbf{gu} & 0 & 0 & 0 & \textbf{58} & 0 & 0 & 0 & 0 & 0 & 0 & 0 & 0 & 0 & 1 & 1 & 1 \\ \cline{2-18} 
 & \textbf{hi} & 0 & 0 & 0 & 0 & \textbf{57} & 0 & 0 & 0 & 0 & 0 & 0 & 1 & 0 & 0.983 & 0.983 & 0.983 \\ \cline{2-18} 
 & \textbf{kn} & 0 & 0 & 0 & 0 & 0 & \textbf{58} & 0 & 0 & 0 & 0 & 0 & 0 & 0 & 1 & 1 & 1 \\ \cline{2-18} 
 & \textbf{ml} & 0 & 0 & 0 & 0 & 1 & 0 & \textbf{56} & 0 & 0 & 0 & 1 & 0 & 0 & 1 & 0.966 & 0.982 \\ \cline{2-18} 
 & \textbf{mn} & 0 & 0 & 0 & 0 & 0 & 0 & 0 & \textbf{58} & 0 & 0 & 0 & 0 & 0 & 0.983 & 1 & 0.991 \\ \cline{2-18} 
 & \textbf{mr} & 0 & 0 & 0 & 0 & 0 & 0 & 0 & 0 & \textbf{56} & 0 & 0 & 0 & 0 & 1 & 1 & 1 \\ \cline{2-18} 
 & \textbf{or} & 0 & 0 & 0 & 0 & 0 & 0 & 0 & 0 & 0 & \textbf{58} & 0 & 0 & 0 & 1 & 1 & 1 \\ \cline{2-18} 
 & \textbf{rj} & 0 & 0 & 0 & 0 & 0 & 0 & 0 & 1 & 0 & 0 & \textbf{57} & 0 & 0 & 0.919 & 0.983 & 0.95 \\ \cline{2-18} 
 & \textbf{ta} & 0 & 0 & 0 & 0 & 0 & 0 & 0 & 0 & 0 & 0 & 4 & \textbf{54} & 0 & 0.964 & 0.931 & 0.947 \\ \cline{2-18} 
 & \textbf{te} & 0 & 0 & 0 & 0 & 0 & 0 & 0 & 0 & 0 & 0 & 0 & 1 & \textbf{57} & 1 & 0.983 & 0.991 \\
    \hline
    \end{tabular}
\end{table}

\label{lastpage}

\end{document}